\documentclass{article}
\usepackage{ijcai21}

\usepackage{times}

\usepackage{soul}
\usepackage{url}
\usepackage[hidelinks]{hyperref}
\usepackage[utf8]{inputenc}
\usepackage[small]{caption}
\usepackage{graphicx}
\usepackage[inference]{semantic}
\usepackage{pifont}
\usepackage{amsmath,amssymb,amsthm}
\usepackage{booktabs}
\usepackage[all]{xy}
\allowdisplaybreaks

\usepackage{misc}
\usepackage{xspace}
\usepackage{xcolor}

\newcommand{\bet}{{\,\mathbin{\bowtie}\,}}
\newcommand{\mycite}[1]{\citeauthor{#1}~\shortcite{#1}}
\newcommand{\inv}[1]{#1^{\smallsmile}}
\newcommand{\anaEquiv}{\sim}
\newtheorem{example}{Example}
\newtheorem{theorem}{Theorem}
\newtheorem{proposition}{Proposition}
\newtheorem{lemma}{Lemma}

\newtheorem{corollary}{Corollary}
\newtheorem{definition}{Definition}
\newcommand{\ana}[4]{#1{\triangleright}#2{::}#3{\triangleright}#4}
\newcommand{\sana}[4]{#1{\blacktriangleright}#2{::}#3{\blacktriangleright}#4}
\newcommand{\ap}[4]{#1{:}#2\,{::}\,#3{:}#4}

\newcommand{\complexity}[1]{}
\newif\ifappendix
\appendixtrue

\title{A Description Logic for Analogical Reasoning}

\author{%
Steven Schockaert \and Yazm\'in  Ib\'a\~nez-Garc\'ia\and
V\'ictor Guti\'errez-Basulto
\affiliations
Cardiff University, UK\\
\emails
\{schockaerts1,ibanezgarciay, gutierrezbasultov\}@cardiff.ac.uk}

\begin{document}

\maketitle

\begin{abstract}
Ontologies formalise how the concepts from a given domain are interrelated. Despite their clear potential as a backbone for explainable AI, existing ontologies tend to be highly incomplete, which acts as a significant barrier to their more widespread adoption.
To mitigate this issue, we present a mechanism to infer plausible missing knowledge, which relies on reasoning by analogy. 
To the best of our knowledge, this is the first paper that studies analogical reasoning within the setting of description logic ontologies. After showing that the standard formalisation of analogical proportion has important limitations in this setting, we introduce an alternative semantics based on bijective mappings between sets of features. We then analyse the properties of analogies under the proposed semantics, and show among others how it enables two plausible inference patterns: rule translation and rule extrapolation. 
\end{abstract}

\section{Introduction}
The last decade has witnessed an increasing interest in methods for automated knowledge base completion. While most work has focused on predicting plausible missing facts in knowledge graphs \cite{NIPS20135071,DBLP:journals/corr/YangYHGD14a,DBLP:journals/jmlr/TrouillonDGWRB17,DBLP:conf/emnlp/BalazevicAH19}, some authors have also looked at the problem of predicting plausible missing rules in ontologies \cite{beltagy2013montague,DBLP:conf/aaai/BouraouiS19}. The underlying principle of these latter approaches is to rely on external knowledge about the similarity structure of concepts, typically in the form of a vector space embedding of concept names. The main idea is that knowledge about concepts can often be extended to concepts with a similar representation in the given vector space. The same principle also lies at the basis of rule-based frameworks with a soft unification mechanism \cite{Medina200443,DBLP:conf/nips/Rocktaschel017}. 
However, these similarity based reasoning methods can clearly only provide us with knowledge that is similar to what is already in the knowledge base. Humans, on the other hand, can also infer plausible knowledge in more creative ways, where a particularly prominent role is played by the idea of reasoning by analogy. This phenomenon has been extensively studied in cognitive science and philosophy \cite{gentner1983structure,hofstadter1995copycat,holyoak1997analogical}, but to the best of our knowledge, the use of analogical reasoning for completing ontologies has not yet been considered\footnote{We note that the term ``analogical reasoning'' has been used in the literature to refer to a form of similarity-based ontology completion \cite{DBLP:conf/semweb/dAmatoFE06}. In contrast, we reserve this term for methods that require drawing parallels between different domains.}

Within Artificial Intelligence, the formalisation of analogical reasoning typically builds on analogical proportions, i.e.\ statements of the form {\it ``$A$ is to $B$ what $C$ is to $D$''} \cite{DBLP:conf/ijcai/BayoudhMD07,DBLP:series/sci/PradeR14a,barbot2019analogy}. A key result in this area has been the development of analogical classifiers, which are based on the principle that whenever the features of four examples are in an analogical proportion, then their class labels should be in an analogical proportion as well \cite{DBLP:conf/ijcai/BayoudhMD07,DBLP:conf/ecai/HugPRS16}. The same principle can be applied to infer plausible concept inclusions, i.e.\ concept inclusions which are not entailed from a given TBox, but which are likely to hold given additional background knowledge that we have about analogical relationships between different concepts. The resulting inference pattern, which we call \emph{rule extrapolation}, is illustrated in the next example.
\begin{example}[Rule extrapolation]\label{exRuleExtrapolation}
Suppose we have an ontology with the following concept inclusions:
\begin{align}
\textit{Young} \sqcap \textit{Cat} &\sqsubseteq \textit{Cute}\label{eqRuleExtrapolation1}\\
\textit{Adult} \sqcap \textit{WildCat} &\sqsubseteq \textit{Dangerous}\label{eqRuleExtrapolation2}\\
\textit{Young} \sqcap \textit{Dog} &\sqsubseteq \textit{Cute}\label{eqRuleExtrapolation3}
\end{align}
Suppose we are furthermore given that ``\emph{Cat} is to \emph{WildCat} what \emph{Dog} is to \emph{Wolf}''. Trivially, we also have that ``\emph{Young} is to \emph{Adult} what \emph{Young} is to \emph{Adult}'' and ``\emph{Cute} is to \emph{Dangerous} what \emph{Cute} is to \emph{Dangerous}''
Using rule extrapolation, we can then infer the following:
\begin{align}\label{eqRuleExtrapolation4}
\textit{Adult} \sqcap \textit{Wolf} &\sqsubseteq \textit{Dangerous}
\end{align}
\end{example}

\noindent Analogies can also be used to infer plausible knowledge by allowing us to transfer knowledge from one domain to another. This is illustrated in the following example.
\begin{example}[Rule translation]\label{exRuleTranslation}
Suppose we are given the following knowledge:
\begin{align}\label{eqRuleTranslation1}
\textit{Program} &\sqsubseteq \exists \textit{specifies}. \textit{Software}
\end{align}
and the fact that ``\emph{Program} is to \emph{Plan} what \emph{Software} is to \emph{Building''.}
Then we can plausibly infer: 
\begin{align}\label{eqRuleTranslation2}
\textit{Plan} &\sqsubseteq \exists \textit{specifies}. \textit{Building}
\end{align}
\end{example}  
\noindent Ontologies often use the same ``templates'' to encode knowledge from different domains (e.g.\ knowledge about different professions). The strategy from Example \ref{exRuleTranslation} then allows us to complete the ontology by introducing additional domains.

Using analogies for identifying plausible missing knowledge is appealing, because they can be learned from text quite effectively. For instance, \mycite{DBLP:journals/coling/Turney06} proposed a method for identifying similarities between pairs of words (i.e.\ analogical proportions) using matrix factorization and ternary co-occurrence statistics, which approximated the performance of an average US college applicant. Moreover, the GPT-3 language model is able to identify analogical word pairs with even higher accuracy \cite{DBLP:conf/nips/BrownMRSKDNSSAA20}. To a more limited extent, some types of analogical relationships can also be obtained from word embeddings \cite{DBLP:conf/naacl/MikolovYZ13}.


The main aim of this paper is to propose a  semantics for modelling analogies within the setting of description logics, which introduces a number of unique challenges (see Section \ref{secAnalogicalConcepts}). We focus in particular on an extension of $\EL_\bot$. We start from the semantics proposed by \mycite{KR2020interpolation}, which extends the $\EL$ semantics by assigning to each individual a set of features. Their aim was to support another form of plausible inference, called interpolation (see Section \ref{sec:back}). We show that having access to these features in the semantics also allows us to formalise analogies. 
\complexity{We study the properties of the resulting semantics and show that checking concept subsumption is \coNP-complete in the considered logic.}

\section{Background}~\label{sec:back}
We start by introducing the Description Logic (DL) $\EL_\bot^{\bowtie}$, which is a straightforward extension of the logic $\EL^{\bowtie}$ that was proposed in \cite{KR2020interpolation} to formalise \emph{rule interpolation}\footnote{We include $\bot$ because disjointness will play an important role in this paper.}. Our approach to analogical reasoning in this paper will  build on $\EL_\bot^{\bowtie}$. Rule interpolation is another inference pattern for obtaining plausible missing concept inclusions in a DL ontology. Interpolation is based on the notion of \emph{betweenness}, where a concept $B$ is said to be between concepts $A$ and $C$ if $B$ has all the natural properties that $A$ and $C$ have in common. In such a case, knowledge that holds for both $A$ and $C$ seems likely to hold for $B$ as well. This inference pattern is illustrated in the next example.
\begin{example}[Rule interpolation]\label{exInterpolation}
Suppose we have the following concept inclusions:
\begin{align}\label{eqExInterpolation1}
\textit{Cat} &\sqsubseteq X &
\textit{Wolf} &\sqsubseteq X
\end{align}
As long as $X$ is a ``natural concept'', it seems plausible that the following concept inclusion also holds:
$$
\textit{Dog} \sqsubseteq X
$$
This is because all the common (natural) properties of Cat and Wolf are also satisfied by Dog. In such a case, we say that the concept Dog is \emph{between} the concepts Cat and Wolf.
\end{example}
\noindent The notion of naturalness plays an important role in most philosophical accounts of induction. Intuitively speaking, a natural concept or property is one that admits inductive inferences. The semantics from \cite{KR2020interpolation} is based on the common view that natural concepts are those which can be characterised as a set of features (i.e.\ a conjunction of elementary properties) \cite{tversky1977features}. 

\smallskip
\noindent\textbf{Syntax.} The logic $\EL_\bot^{\bowtie}$  extends the standard DL $\EL_\bot$ with \emph{in-between concepts} of the form    
$C \bet D $, describing the set of objects that are between the concepts $C$ and $D$. 
Further, $\EL_\bot^{\bowtie}$ includes an infinite set of \emph{natural concept names}. More precisely, consider countably infinite
but disjoint sets of \emph{concept names} $\mn{N_C}$ and \emph{role names} $\mn{N_R}$, where $\mn{N_C}$ contains a distinguished  infinite set of \emph{natural concept names} $\mn{N^{Nat}_C}$.  
The syntax of \emph{$\EL_\bot^\bet$ concepts $C,D$} is  defined  by the following grammar, where $A \in \mn{N_C}$, 
$A' \in \mn{N^{Nat}_C}$ and $r \in \mn{N_R}$:
\begin{align*} 
C,D &:= \top \mid \bot \mid A \mid C \sqcap D \mid \exists r. C \mid  N     \\
N,N'  &:= A' \mid N \sqcap N' \mid N \bet N'
\end{align*}
Concepts of the form $N,N'$ are called \emph{natural concepts}.
An \emph{$\EL_\bot^\bet$ TBox} is a finite set of 
concept inclusions $C\sqsubseteq D$, where  $C,D$ are $\EL_{\bot}^\bet$ concepts.

\smallskip
\noindent\textbf{Semantics.} The semantics of $\EL_\bot^{\bowtie}$ is given in terms of feature-enriched interpretations, which extend standard first-order interpretations by also specifying a mapping $\pi$ from individuals to sets of 
\emph{features}. Formally, a  \emph{feature-enriched} interpretation is a tuple $\Imf= ( \Imc,  \Fmc, \pi)$ in which $\Imc = (\Delta^\Imc, \cdot^\Imc)$ is a classical DL interpretation, $\mathcal{F}$ is a finite set of features, and $\pi: \Delta^{\Imc}\rightarrow 2^{\mathcal{F}}$, such that the following hold:
\begin{enumerate}
\item For each $d\in\Delta^{\Imc}$ it holds that $\pi(d)\subset \mathcal{F}$;
\item for each $F\subset \mathcal{F}$ there exists some individual $d\in\Delta^{\Imc}$ such that $\pi(d)=F$.
\end{enumerate}
For an $\EL_\bot^{\bowtie}$ concept $C$, $C^\Imf$ is defined as a pair $\langle C^\Imc, \varphi(C)\rangle$ where $C^\Imc \subseteq \Delta^\Imc$ and $\varphi(C)$ is the set of all features associated with a concept $C$, defined as: 
$$
\varphi(C) = \bigcap\{\pi(d) \mid d\in C^\Imc\}
$$
Intuitively, the set of features $\varphi(C)$ describes the concept $C$ at a finer-grained level that what may be possible in the language. This makes it possible to capture knowledge about what different concepts have in common, which is needed in $\EL_\bot^{\bowtie}$ to model the semantics of in-between concepts. 

For a standard  $\EL_\bot$ concept $C$, $C ^\Imc$ is defined as usual~\cite{DBLP:books/daglib/0041477}. For in-between concepts, $\cdot^\Imc$ is defined as follows.
$$
(N \bet  N' )^\Imc  = \{d \in \Delta^\Imc \mid   \varphi(N)\cap \varphi(N') \subseteq \pi(d)\}.
$$
Intuitively, $(N \bet N')^\Imc$ contains all elements from the domain that have all the features that are common to both $N$ and $N'$.
A feature-enriched interpretation $\Imf  = (\Imc, \Fmc, \pi )$ \emph{satisfies} a concept inclusion $C\sqsubseteq D$ if $C^\Imc \subseteq D^\Imc$. \Imf is a \emph{model} of an $\EL^\bet$ TBox $\Tmc$ if it satisfies all CIs in \Tmc 
    and for every natural concept  $N$ in $\Tmc$, it holds that
\begin{align}\label{eqConditionNatural}
N^\Imc = \{d \in \Delta^\Imc \,|\, \varphi(N) \subseteq \pi(d)\}
\end{align}
i.e.\  $N$ is fully specified by its features. It is easy to verify that \eqref{eqConditionNatural} is satisfied for a complex natural concept, as soon as it is satisfied for its constituent natural concept names. 
%
%
A concept    $C$ \emph{is satisfiable  w.r.t.\ a TBox \Tmc},  if there is a model $\Imf$ of \Tmc such that $C^\Imf \neq \langle \emptyset, \mathcal F  \rangle$. 

\smallskip
The purpose of introducing features in the semantics of $\EL_\bot^{\bowtie}$ is to make explicit what different concepts have in common, and to use this as the basis for enabling a particular kind of inductive inference (i.e.\ interpolation). For instance, in the case of Example \ref{exInterpolation}, if the concept inclusions in \eqref{eqExInterpolation1} are complemented with $\textit{Dog}\sqsubseteq \textit{Cat}\bowtie\textit{Wolf}$, it can be verified that $\textit{Dog}\sqsubseteq X$ can indeed be inferred (assuming all concept names are natural). In applications, knowledge about in-between concepts would typically be induced from a vector space embedding. See, for instance, \cite{DBLP:conf/aaai/BouraouiS19} for a practical application of rule interpolation based on pre-trained word embeddings. In this paper, we will build on the feature-enriched semantics to encode correspondences between analogous domains.

\section{Analogical Concepts}\label{secAnalogicalConcepts}
Analogical proportions are a central notion in the formalisation of analogical reasoning, going back to Aristotle (see \cite{barbot2019analogy} for a historical perspective). While they can be defined more generally, here we will focus on analogical proportions between sets. In particular, the sets $S_1,S_2,S_3,S_4$ are said to be in an analogical proportion, denoted as $\ap{S_1}{S_2}{S_3}{S_4}$ if $S_1$ and $S_2$ differ in the same way that $S_3$ and $S_4$ differ. Formally, $\ap{S_1}{S_2}{S_3}{S_4}$ is satisfied if:
\begin{align}
S_1 \setminus S_2 &= S_3 \setminus S_4 &
S_2 \setminus S_1 &= S_4 \setminus S_3\label{eqDefAP}
\end{align}
Some key properties of analogical proportions are as follows:
\begin{description}
\item[Reflexivity] $\ap{A}{B}{A}{B}$
\item[Symmetry] $(\ap{A}{B}{C}{D}) \Leftrightarrow (\ap{C}{D}{A}{B})$
\item[Exchange of means] $(\ap{A}{B}{C}{D}) \Leftrightarrow (\ap{A}{C}{B}{D})$
\item[S-transitivity] $(\ap{A}{B}{C}{D}) {\wedge} (\ap{A}{B}{E}{F}) {\Rightarrow} (\ap{C}{D}{E}{F})$
\item[C-transitivity] $(\ap{A}{C}{D}{B}) {\wedge} (\ap{A}{E}{F}{B}) {\Rightarrow} (\ap{C}{E}{F}{D})$
\end{description}

\subsection{Analogical Proportions between DL Concepts}
In this paper, we are concerned with analogies between description logic concepts. We can define analogical proportions between $\EL_\bot^{\bowtie}$ concepts $A,B,C,D$ as $\ap{A^{\Imc}}{B^{\Imc}}{C^{\Imc}}{D^{\Imc}}$ or as $\ap{\varphi(A)}{\varphi(B)}{\varphi(C)}{\varphi(D)}$. In general these two expressions are not equivalent, and there are advantages in requiring that both of them are satisfied at the same time, which has been studied in detail in \cite{barbot2019analogy} within the setting of Formal Concept Analysis. However, if $A,B,C,D$ are natural, then $\ap{A^{\Imc}}{B^{\Imc}}{C^{\Imc}}{D^{\Imc}}$ and $\ap{\varphi(A)}{\varphi(B)}{\varphi(C)}{\varphi(D)}$ are equivalent. Moreover, for natural concepts,  the semantic constraint $\ap{\varphi(A)}{\varphi(B)}{\varphi(C)}{\varphi(D)}$ can be modelled syntactically in  $\EL_\bot^{\bowtie}$. Indeed it holds that $\ap{\varphi(A)}{\varphi(B)}{\varphi(C)}{\varphi(D)}$ is satisfied iff the following conditions are satisfied\footnote{This follows immediately from the characterisation of Boolean analogical proportions in terms of conjunction and disjunction; see  \cite{DBLP:journals/lu/PradeR13} for details.}:
\begin{align}
\varphi(A)\cap\varphi(D) &= \varphi(B)\cap\varphi(C) \label{eqAlternativeAP1} \\
\varphi(A)\cup\varphi(D) &= \varphi(B)\cup\varphi(C)\label{eqAlternativeAP2}
\end{align}
The analogical proportion
$\ap{\varphi(A)}{\varphi(B)}{\varphi(C)}{\varphi(D)}$ is thus satisfied if the following concept inclusions are satisfied:
\begin{align*}
A \sqcap D &\sqsubseteq B \sqcap C & B \sqcap C &\sqsubseteq A \sqcap D\\
A \bowtie D &\sqsubseteq B \bowtie C & B \bowtie C &\sqsubseteq A \bowtie D
\end{align*}
In the following, we will write $\ap{A}{B}{C}{D}$ as an abbreviation for these four concept inclusions.
The fact that we can model $\ap{A}{B}{C}{D}$ within $\EL_\bot^{\bowtie}$ is an advantage, but as we will see below, modelling analogies between DL concepts in this way has a number of important limitations. 

\subsection{Desiderata for Modelling Analogies in DLs}\label{secDesiderata}
Our motivation for studying analogies is to enable plausible inferences. A clear requirement is thus that we want some form of the \emph{rule extrapolation} and \emph{rule translation} inference patterns, as illustrated in Examples \ref{exRuleExtrapolation} and \ref{exRuleTranslation}, to be satisfied. Another important requirement comes from the fact that we usually only have access to information about analogical relationships between concept names (e.g.\ obtained from a language model). However, the aforementioned inference pattern may rely on analogical relationships between complex concepts. To enable non-trivial inferences in practice, it is thus important that analogies between concept names can be \emph{lifted} to analogies between complex concepts.

Let us now consider the suitability of analogical proportions, in light of these desiderata. First, rule translation is satisfied for analogical proportions.
\begin{proposition}\label{propIRFBSA1}
Let $\Imf= ( \Imc,  \mathcal{F}, \pi)$ be a feature-enriched interpretation, and let $A_1,A_2,B_1,B_2$ be natural concepts in $\Imf$. If $\ap{\varphi(A_1)}{\varphi(A_2)}{\varphi(B_1)}{\varphi(B_2)}$ holds and $\Imf$ satisfies $A_1 \sqsubseteq B_1$, then $\Imf$ also satisfies $A_2 \sqsubseteq B_2$.
\end{proposition}
\noindent However, rule extrapolation is not valid for analogical proportions. 
Furthermore, analogical proportions between concept names cannot be lifted to complex concepts. For instance, from $\ap{A_1}{B_1}{C_1}{D_1}$ and $\ap{A_2}{B_2}{C_2}{D_2}$, in general we cannot infer $\ap{(A_1\sqcap A_2)}{(B_1\sqcap B_2)}{(C_1\sqcap C_2)}{(D_1\sqcap D_2)}$. Counterexamples  are provided in the appendix.

\section{The Logic $\EL_{\bot}^{\textit{ana}}$}
Given the limitations of analogical proportions that were highlighted in Section \ref{secDesiderata}, we propose an alternative approach for modelling analogies between description logic concepts. This approach is based on the common view that analogies are mappings from one domain into another, which lies among others at the basis of the seminal Structure Mapping framework \cite{gentner1983structure}. In particular, we propose the logic $\EL_{\bot}^{\textit{ana}}$, which extends $\EL_{\bot}^{\bowtie}$ with two novel elements: analogy assertions and intra-domain roles. 
Analogy assertions are similar to analogical proportions, in that they encode a relationship of the form ``$A$ is to $B$ what $C$ is to $D$'', but their semantics is defined in terms of mappings between different domains, where domains will be identified with sets of features.
Intuitively, intra-domain roles are roles which preserve the structure of analogous domains.

\subsection{Syntax}
Let $\mn{N_C}$, $\mn{N^{Nat}_C}$ and $\mn{N_R}$ be defined as before.
We assume that $\mn{N_R}$ contains an infinite set $\mn{N^{\textit{Int}}_R}$ of distinguished  \emph{intra-domain role names}. 
The syntax of $\EL^{\textit{ana}}_{\bot}$ concepts $C,D$ is defined  by the following grammar, where $A \in \mn{N_C}$, 
$A' \in \mn{N^{Nat}_C}$, $r \in \mn{N_R}$ and $r' \in \mn{N_R^{\textit{Int}}}$:
\begin{align*} 
C,D &:= \top \mid \bot \mid A \mid C \sqcap D \mid \exists r. C \mid  N     \\
N,N'  &:= A' \mid N \sqcap N' \mid N \bet N' \mid \exists r'. N
\end{align*}
$\EL_{\bot}^{\textit{ana}}$ concepts extend $\EL_{\bot}^{\bowtie}$  concepts by allowing existential restrictions over intra-domain roles as natural concepts. 
An $\EL^{\textit{ana}}_{\bot}$ TBox is a finite set containing two types of assertions: (i) $\EL^{\textit{ana}}_{\bot}$ concept inclusions, and 
 (ii) \emph{analogy assertions} of the form $\ana{C_1}{D_1}{C_2}{D_2}$, where $C_1,C_2,D_1,D_2$ are natural $\EL^{\textit{ana}}_{\bot}$ concepts.


\subsection{Semantics}
Analogies intuitively involve transferring knowledge from one domain to another domain\footnote{In this paper, we use the term domain to refer to a particular thematic area. This should not be confused with the set $\Delta^\Imc$, which is often referred to as the domain of the interpretation $\Imc$.}, e.g.\ from software engineering to architecture in the case of Example \ref{exRuleTranslation}. In our framework, these domains will be associated with subsets of $\mathcal{F}$. In particular, we will require that interpretations specify a partition $[\mathcal{F}_1,...,\mathcal{F}_k]$ of $\mathcal{F}$, defining the different domains of interest. Some of the partition classes will furthermore be viewed as being analogous, in the sense that there is some kind of structure-preserving mapping between them. 

Another extension of the feature-enriched semantics is aimed at improving how disjointess can be modelled.
In the semantics from \cite{KR2020interpolation}, no individual $d\in\Delta^{\Imc}$ is allowed to have all the features from $\mathcal{F}$, but all proper subsets $F\subset \mathcal{F}$ are witnessed in the sense that there is some $d$ such that $\pi(d)=F$. This limits how disjoint concepts can be modelled.{ For instance, $B$ cannot be satisfied w.r.t.\ $\{B\sqsubseteq A\bowtie C, A\sqcap B \sqsubseteq \bot,A\sqcap C \sqsubseteq \bot, B\sqcap C \sqsubseteq \bot\}$ using a feature-enriched interpretation.} Disjointness will play an important role in our semantics, as concepts from different domains will be required to be disjoint (see below). For this reason, we extend the feature-enriched semantics with sets of \emph{forbidden feature combinations}. In particular, interpretations will specify a set $\mathcal{X}\in 2^{\mathcal{F}}$ such that for $X\in \mathcal{X}$, it holds that no individual can have all the features from $X$. For the ease of presentation, we write $\mathcal{C}$ for the set of all \emph{consistent sets of features}, i.e.\ $F\in \mathcal{C}$ iff $F\not\supseteq X$ for all $X\in\mathcal{X}$. We also write $\mathcal{C}^i$ for the restriction of $ \mathcal{C}$ to subsets of $\mathcal{F}_i$.

\begin{definition}\label{defDomainInterpretation}
Let $[\mathcal{F}_1,...,\mathcal{F}_k]$ be a partition of a non-empty finite set $\Fmc$. We call $\Imf= ( \Imc,  [\mathcal{F}_1,...,\mathcal{F}_k], \mathcal{X}, \pi, \anaEquiv, \mathcal{S})$ a \emph{domain constrained interpretation} if $\Imc=(\Delta^\Imc,.^{\Imc})$ is a classical DL interpretation,  $\mathcal{X} \subseteq 2^{\Fmc}$, $\Fmc \in \mathcal{X}$, $\pi:\Delta^{\Imc}\rightarrow 2^{\Fmc}$, 
$\anaEquiv$  is an equivalence relation over $\{1,...,k\}$ and $S=\{\sigma_{(s,t)}\,|\, (s,t)\in \anaEquiv\}$, with each $\sigma_{s,t}$ a bijection from $\mathcal{F}_s$ to $\mathcal{F}_t$, and we have:
\begin{enumerate}
\item for every $d\in\Delta^\Imc$ and $X\in \mathcal{X}$ it holds that $X \not\subseteq \pi(d)$;
\item if $X\in \mathcal{C}$ then $\pi(d)=X$ for some $d\in\Delta^\Imc$;
\item we have $\sigma_{(s,t)}^{-1} =\sigma_{(t,s)}$ and $\sigma_{(t,u)}\circ \sigma_{(s,t)}=\sigma_{(s,u)}$ for any $(s,t),(t,u)\in \anaEquiv$;
\item for $F\in \mathcal{C}^i$ and $(i,j)\in\anaEquiv$, we have $\{\sigma_{(i,j)}(f)\,|\, f\in F \} \in\mathcal{C}$;
\item if $f\in \mathcal{F}_i \text{ and } g\in\mathcal{F}_j \text{ then } \{f,g\}\in\mathcal{X}$, for all $(i,j)\in\anaEquiv$ with $i\neq j$.
\end{enumerate}
\end{definition}

\noindent Intuitively, each of the sets $\mathcal{F}_i$ corresponds to a different domain. If $(s,t)\in\anaEquiv$, it means that there is an analogy between the \emph{source domain} $\mathcal{F}_s$ and \emph{the target domain} $\mathcal{F}_t$. In that case, there is a one-to-one mapping $\sigma_{(s,t)}$ between the features from $\mathcal{F}_s$ and those from $\mathcal{F}_t$. The first two conditions from Definition \ref{defDomainInterpretation} capture the fact that a set of features $X\subseteq \mathcal{F}$ is witnessed by some individual iff it is consistent, i.e.\ $X\in \mathcal{C}$. The third condition ensures that the mappings $\sigma_{(i,j)}$ can be composed and reversed. The fourth condition encodes that the mapping $\sigma_{(i,j)}$ maps consistent feature combinations to consistent feature combinations. 
This is a natural requirement, given the intuition that analogous domains should have the same structure. 
The last condition captures the requirement that individuals cannot have features from two analogous domains. While analogies are normally indeed defined between distinct domains, the reader may wonder at this point whether this restriction is necessary. We will come back to this question in Section \ref{secWeakSemantics}.

\paragraph{Domain Translations}
Before presenting the semantics of analogy assertions, we first study how the bijections $\sigma_{(i,j)}$ can be combined to define mappings between the sets of features $\varphi(C),\varphi(D)$ associated with two concepts.
First, we define a \emph{domain assignment mapping} $\delta$, which maps each concept $C$ to the set of domains on which it depends:
$$
\delta(C) = \{i \mid \mathcal{F}_i\cap \varphi(C)\neq \emptyset\}
$$
Next, we extend the mappings $\sigma_{(i,j)}$ to mappings between sets of domains. Let $U=\{(s_1,t_1), \ldots ,(s_l,t_l)\}$, with $s_1, \ldots ,s_l$ all distinct and $t_1, \ldots ,t_l$ all distinct. The mapping $\sigma_U$ is defined as follows: 
\begin{align*}
\sigma_U(f) =
\begin{cases}
\sigma_{(s_i,t_i)}(f) & \text{if $f\in \mathcal{F}_{s_i}$}\\
f & \text{otherwise}
\end{cases}
\end{align*}
We call $\sigma_U$ a \emph{domain translation}, and write $\textit{src}(U)$ for the set $\{s_1,\ldots ,s_l\}$ of source domains and $\textit{tgt}(U)$ for the set $\{t_1,\ldots ,t_l\}$ of target domains. 
The source domains need to be distinct to ensure that the domain translation is uniquely defined.
Target domains are required to be distinct to allow domain translations to be reversible. For the ease of presentation, we will write $\sigma_U(F)$ to denote the set $\{\sigma_U(f) \,|\, f\in F\}$. 

For a pair of concepts $C$ and $D$, we write $\mu(C,D)$ for the set of domain translations $\sigma_U$ such that:
\begin{align}
\varphi(D)&= \sigma_U(\varphi(C))\label{eqDefSigmaAP}\\
\textit{src}(U)&\subseteq \delta(C)\label{eqNoRedundantMappings}\\
\textit{tgt}(U)\cap (\delta(C) \setminus \textit{src}(U))&=\emptyset\label{eqTargetSourceDisjointness}
\end{align}
The first condition states that the domain translations in $\mu(C,D)$ essentially ``translate'' the concept $C$ to the concept $D$. The second condition ensures that $\mu(C,D)$ contains minimal domain translations only, in the sense that $U$ should not contain any redundant pairs. The third condition is needed to ensure that domain translations are reversible. To see why this is needed, let $\varphi(C)=\{f_1,g_2\}$, $\varphi(D)=\{g_1,g_2\}$, $\mathcal{F}_1=\{f_1,f_2\}$, $\mathcal{F}_2=\{g_1, g_2\}$, $\sigma_{(1,2)}(f_i)=g_i$. Then $\varphi(D)=\sigma_{(1,2)}(\varphi(C))$, but there is no domain translation $\sigma_{U}$ s.t.\ $\varphi(C)=\sigma_{U}(\varphi(D))$. As the following result shows, imposing \eqref{eqTargetSourceDisjointness} is enough to ensure reversibility. 
\begin{proposition}\label{propDomainTranslationsReversed}
Let $U$ be a domain translation, and let $\inv{U}=\{(t,s) \,|\, (s,t)\in U\}$. It holds that $\sigma^{-1}_{U}=\sigma_{\inv{U}}$.
\end{proposition}
\noindent As the next result shows, the composition of two domain translations is also a valid domain translation. In particular, if there is some domain translation $\sigma_{U}$ that maps $C$ to $D$ and some domain translation $\sigma_{V}$ that maps $D$ to $E$, then $\sigma_{U}$ and $\sigma_{V}$ can be composed to define a domain translation from $C$ to $E$. Moreover, in such a case, any domain translation from $C$ to $E$ can be defined as such a composition.
\begin{proposition}\label{propMuComposition}
If $\mu(C,D)\neq \emptyset$ and $\mu(D,E)\neq \emptyset$ we have:
$$
\mu(C,E) = \{\sigma_{U\oplus V} \,|\, \sigma_{U}\in\mu(C,D), \sigma_{V}\in\mu(D,E)\}
$$
where 
\begin{align*}
U\oplus V = &\{(i,k) \,|\, (i,j)\in U, (j,k)\in V, i\neq k\}\\
 &\cup \{(i,j) \,|\, (i,j)\in U, j\notin \textit{src}(V)\}\\
 &\cup \{(j,k) \,|\, (j,k)\in V, j\notin \textit{tgt}(U)\}
\end{align*}
\end{proposition}
\paragraph{Semantics of Intra-Domain Roles}
We will need to put additional constraints on the interpretation of a role $r$ to be able to infer $\ana{(\exists r.A)}{(\exists r.B)}{(\exists r.C)}{(\exists r.D)}$ or $\ana{A}{B}{(\exists r.C)}{(\exists r.D)}$ from $\ana{A}{B}{C}{D}$. To allow such lifting of analogy assertions, we will associate with each intra-domain role $r$ a mapping $\kappa_r$ between sets of features, which satisfies a number of conditions. In particular, we introduce the following notion of intra-domain relation.
\begin{definition}\label{defIntraDomainRole}
Let $\Imf= ( \Imc,  [\mathcal{F}_1,...,\mathcal{F}_k], \mathcal{X}, \pi, \anaEquiv, \mathcal{S})$ be a domain constrained interpretation and let $r\in \mn{N_R}$. We say that $r$ is  \emph{interpreted as an intra-domain relation} if for every concept $C$,  we have
$(\exists r.C)^{\Imc}=\{d\in \Delta^{\Imc} \,|\, \pi(d)\supseteq \kappa_r(\varphi(C)\}$, for a mapping $\kappa_r$ satisfying:
\begin{enumerate}
\item $\kappa_r(F) = \kappa_r(F\cap \mathcal{F}_1)\cup ...\cup \kappa_r(F\cap \mathcal{F}_k)$, for all $F\in \mathcal{C}$;
\item $\kappa_r(F)\subseteq \mathcal{F}_i$, for all $i\in \{1,...,k\}$ and $F\in \mathcal{C}^i$;
\item $\kappa_r (\sigma_{\{(i,j)\}}(F)) = \sigma_{\{(i,j)\}}( \kappa_r(F))$, for all $(i,j)\in \anaEquiv$ and $F\in\mathcal{C}^i$;
\item $\kappa_r(F)\neq\emptyset$, for all $i\in \{1,...,k\}$ and $F\in\mathcal{C}^i\setminus\{\emptyset\}$.
\end{enumerate}
\end{definition}
\noindent The first two conditions in Definition \ref{defIntraDomainRole} state that the features in $\kappa_r(F)$ are  determined per domain. The third condition captures the intuition that analogous domains should have the same structure. The last condition essentially encodes that whenever $C$ depends on some domain $i$  then $\exists r.C$ should also depend on domain $i$, i.e.\ if $\varphi(C)$ contains at least one feature from $\mathcal{F}_i$ then the same should be true for $\kappa_r(\varphi(C))$. Note that if $r$ is interpreted as an intra-domain relation and $C$ is a natural concept, then $\exists r.C$ is a natural concept, whose features are determined by the features in $\varphi(C)$. We then have $\varphi(\exists r.C) = \kappa_r(\varphi(C))$.

The semantics of $\EL_{\bot}^{\textit{ana}}$ concepts can now be defined similarly to  Section~\ref{sec:back}, but we additionally  require that every $r\in \mn{N_R^{\textit{Int}}}$ is interpreted as an intra-domain relation.

\paragraph{Semantics of  TBoxes}

We now define the semantics of $\EL^{\textit{ana}}_{\bot}$ TBoxes. We start with that of analogy assertions.
 We say that \emph{a domain constrained interpretation $\Imf$ satisfies the analogy assertion $\ana{C_1}{C_2}{D_1}{D_2}$} if:
\begin{align}\label{eqWeakSemantics}
\mu(C_1,C_2)\cap\mu(D_1,D_2)&\neq \emptyset 
\end{align}
Clearly, $\ana{C_1}{C_2}{D_1}{D_2}$ is equivalent to $\ana{D_1}{D_2}{C_1}{C_2}$. One may wonder whether \eqref{eqWeakSemantics} is sufficient, i.e.\ whether we should not require $\mu(C_1,C_2)=\mu(D_1,D_2)$. However, as the following result shows, for non-empty concepts, $\mu(C_1,C_2)$ and $\mu(D_1,D_2)$ can have at most one element.
\begin{proposition}\label{propStrictUniqueDomainMapping}
Let $\Imf= ( \Imc,  \mathcal{F}, \mathcal{X}, \pi, \anaEquiv, \mathcal{S})$ be a domain-constrained interpretation. If $C^{\Imc}\neq \emptyset$, $D^{\Imc}\neq \emptyset$ and $\mu(C,D)\neq \emptyset$, then $|\mu(C,D)|=1$.
\end{proposition}

\noindent We define the semantics of $\EL^{\textit{ana}}_{\bot}$ TBoxes similarly to Section~\ref{sec:back}, but now including analogy assertions. A domain constrained interpretation \emph{\Imf is a model of an $\EL^{\textit{ana}}_{\bot}$ TBox \Tmc} if \Imf satisfies all CIs and analogy assertions in \Tmc; 
every natural concept $N \in \Tmc$ is fully specified by its features; and every intra-domain role is interpreted as an intra-domain relation. For a TBox $\Tmc$ and CI or analogy assertion $\phi$ we write $\Tmc\models \phi$ to denote that every model of $\Tmc$ satisfies $\phi$. If $\Tmc$ is a singleton of the form $\{\psi\}$, we also write this as $\psi\models\phi$.

\subsection{Properties of Analogy Assertions}\label{secPropertiesAnaAssertions}
\paragraph{Basic Properties}
Before returning to the desiderata from Section \ref{secDesiderata}, we briefly look at the properties of analogical proportions that were listed in Section \ref{secAnalogicalConcepts}. First, reflexivity is trivially satisfied. The symmetry property also holds for analogy assertions, thanks to
the reversibility of domain translations. 
\begin{proposition}\label{propAnaReversible}
We have $\ana{C_1}{C_2}{D_1}{D_2} \models \ana{C_2}{C_1}{D_2}{D_1}$.
\end{proposition}
\noindent Exchange of means is not satisfied. This can easily be seen from the fact that whenever $\ana{C_1}{C_2}{D_1}{D_2}$ is satisfied, we have $|\varphi(C_1)|=|\varphi(C_2)|$ and $|\varphi(D_1)|=|\varphi(D_2)|$ but not necessarily $|\varphi(C_1)|=|\varphi(D_1)|$. As a result of this, there are two variants of S-transitivity that can be considered. As the next proposition shows, both of these variants are satisfied.
\begin{proposition}\label{propTransitivity}
It holds that:
\begin{align}
\{\ana{C_1}{C_2}{D_1}{D_2},\ana{D_1}{D_2}{E_1}{E_2}\} &{\models} \ana{C_1}{C_2}{E_1}{E_2}\label{eqSTransitivityA}\\
\{\ana{C_1}{C_2}{D_1}{D_2},\ana{C_2}{C_3}{D_2}{D_3}\} &{\models} \ana{C_1}{C_3}{D_1}{D_3}\label{eqSTransitivityB}
\end{align}
\end{proposition}
\noindent We also have that $C$-transitivity is satisfied. 
\begin{proposition}\label{propCTransitivity}
It holds that:
\begin{align*}
\{\ana{C_1}{D_1}{D_2}{C_2},\ana{C_1}{E_1}{E_2}{C_2}\} &\models \ana{D_1}{E_1}{E_2}{D_2}
\end{align*}
\end{proposition}

\paragraph{Lifting analogy assertions}
As the next two results show, analogy assertions can indeed be lifted to (non-empty) conjunctions and existentially quantified concepts.

\begin{proposition}\label{propLiftingConjunction}
Let $\Imf=( \Imc,  [\mathcal{F}_1,...,\mathcal{F}_k], \mathcal{X}, \pi, \anaEquiv, \mathcal{S})$ be a domain-constrained interpretation satisfying $(C_i {\sqcap} D_i)^{\Imc}\neq \emptyset$ for $i{\in}\{1 .. 4\}$,
$\ana{C_1}{C_2}{C_3}{C_4}$ and $\ana{D_1}{D_2}{D_3}{D_4}$.
Then $\Imf$ also satisfies
$\ana{(C_1 \sqcap D_1)}{(C_2 \sqcap D_2)}{(C_3 \sqcap D_3)}{(C_4 \sqcap D_4)}$.
\end{proposition}

\begin{proposition}\label{propLiftingExistential}
Let $r$ be an intra-domain role. It holds that:
\begin{align}
\ana{C}{D}{E}{F} &\models \ana{(\exists r.C)}{(\exists r.D)}{(\exists r.E)}{(\exists r.F)}\label{eqLiftingExistential1}\\
\ana{C}{D}{E}{F} &\models \ana{C}{D}{(\exists r.E)}{(\exists r.F)}\label{eqLiftingExistential2}
\end{align}
\end{proposition}

\paragraph{Analogy Based Inference Patterns}
We now return to the two considered analogy based inference patterns: rule translation and rule extrapolation. First, similar as for analogical proportions, we find that rule translation is supported.
\begin{proposition}\label{propSubsumptionTranslation1}
Let $\Imf$ be a domain constrained interpretation. If $\Imf$ satisfies $\{\ana{C_1}{D_1}{C_2}{D_2}, C_1 \sqsubseteq C_2\}$ then $\Imf$ also satisfies $D_1 \sqsubseteq D_2$.
\end{proposition}
\begin{example}
Suppose $\ana{\textit{Program}\,}{\textit{Plan}}{\textit{Software}\,}{\textit{Building}}$ holds and assume that $\textit{specifies}$ is an intra-domain role. Using Proposition \ref{propLiftingExistential} we can then infer:  $$
\ana{\textit{Program}\,}{\textit{Plan}}{(\exists\textit{specifies}.\textit{Software})}{(\exists\textit{specifies}.\textit{Building})}
$$
If we are additionally given that the concept inclusion \eqref{eqRuleTranslation1} is satisfied,  we can infer \eqref{eqRuleTranslation2} using Proposition \ref{propSubsumptionTranslation1}.
\end{example}
\noindent A version of rule extrapolation is also supported.
\begin{proposition}\label{propSubsumptionExtrapolation1}
Let $\Imf= ( \Imc,  [\mathcal{F}_1,...,\mathcal{F}_k], \mathcal{X}, \pi, \anaEquiv, \mathcal{S})$ be a domain constrained interpretation. Suppose that $C_1^{\Imc}\neq \emptyset$ and that $\Imf$ satisfies the following TBox:
\begin{align*}
\Tmc=\{&\ana{C_1}{C_2}{C_3}{C_4},\ana{D_1}{D_2}{D_3}{D_4},\\
&\ana{D_1}{D_3}{D_2}{D_4},C_1 \sqsubseteq D_1, C_2  \sqsubseteq D_2, C_3 \sqsubseteq D_3\}
\end{align*}
Then $\Imf$ also satisfies the assertion $C_4 \sqsubseteq D_4$.
\end{proposition}
\begin{example}
Suppose the concept inclusions \eqref{eqRuleExtrapolation1}--\eqref{eqRuleExtrapolation3} are satisfied, as well as the following analogy assertions\footnote{Note that $\ana{\textit{Cute } }{ \textit{ Cute}}{\textit{Dangerous } }{ \textit{ Dangerous}}$ is trivially satisfied, since $\sigma_{\emptyset}\in \mu(\textit{Cute},\textit{Cute})\cap \mu(\textit{Dangerous},\textit{Dangerous})$.}:
\begin{align*}
\ana{\textit{Young}\,}{\textit{Adult}&}{\textit{Young}\,}{\textit{Adult}}\\ \ana{\textit{Cat}\,}{\textit{WildCat}&}{\textit{Dog}\,}{\textit{Wolf}}\\ \ana{\textit{Cute}\,}{\textit{Dangerous}&}{\textit{Cute}\,}{\textit{Dangerous}}\\ \ana{\textit{Cute}\,}{\textit{Cute}&}{\textit{Dangerous}\,}{\textit{Dangerous}}
\end{align*}
Assuming the intersections involved are all non-empty, using Proposition \ref{propLiftingConjunction} we can infer
$
\ana{(\textit{Young}\sqcap\textit{Cat})}{(\textit{Adult}\sqcap \textit{WildCat})}{(\textit{Young}\sqcap \textit{Dog})}{(\textit{Adult}\sqcap \textit{Wolf})}
$. Finally, using Proposition \ref{propSubsumptionExtrapolation1} we can infer that \eqref{eqRuleExtrapolation4} holds. 
\end{example}
\noindent Note that both $\ana{D_1}{D_2}{D_3}{D_4}$ and $\ana{D_1}{D_3}{D_2}{D_4}$ are required for the above proposition to hold (see the appendix for a counterexample that shows this). For analogical proportions, adding both conditions makes no difference, as $\ap{D_1}{D_2}{D_3}{D_4}$ and $\ap{D_1}{D_3}{D_2}{D_4}$ are equivalent.


\subsection{Alternative Semantics}\label{secWeakSemantics}
As shown in Section \ref{secPropertiesAnaAssertions}, the proposed semantics for analogy assertions satisfies the main desiderata from Section \ref{secDesiderata}. We may wonder, however, whether all aspects of the semantics are necessary for this to hold. We return in particular to Condition 5 from Definition \ref{defDomainInterpretation}, which is perhaps the most restrictive condition. In particular, let us define the notion of \emph{weak domain constrained interpretation} as a tuple $( \Imc,  [\mathcal{F}_1,...,\mathcal{F}_k], \mathcal{X}, \pi, \anaEquiv, \mathcal{S})$ that satisfies all conditions from Definition \ref{defDomainInterpretation}, apart from Condition 5. One immediate consequence of dropping Condition 5 is that Proposition \ref{propStrictUniqueDomainMapping} is no longer valid. As a result, in addition to analogy assertions of the form $\ana{A_1}{A_2}{B_1}{B_2}$, with the semantics defined in \eqref{eqWeakSemantics}, we can now also consider \emph{strong analogy assertions}, denoted as $\sana{A_1}{A_2}{B_1}{B_2}$, which are satisfied if \eqref{eqWeakSemantics} is satisfied and moreover:
$$
\mu(A_1,A_2)=\mu(B_1,B_2)
$$
Under this weak semantics, $C$-transitivity (i.e.\ Proposition \ref{propCTransitivity}) is no longer satisfied, neither for the standard analogy assertions nor for strong analogy assertions (see the appendix for counterexamples). Regarding $S$-transitivity, the variant in \eqref{eqSTransitivityB} remains valid and is furthermore also valid for strong analogy assertions. However, the variant in \eqref{eqSTransitivityA} is no longer valid for standard analogy assertions, although it is satisfied for strong analogy assertions. In contrast, Proposition \ref{propLiftingExistential} remains valid for standard analogy assertions, but it is not satisfied for strong analogy assertions. Proposition \ref{propLiftingConjunction} is neither satisfied for standard analogy assertions nor for strong analogy assertions. Finally, in terms of inference patterns, Proposition \ref{propSubsumptionTranslation1} remains valid, but Proposition \ref{propSubsumptionExtrapolation1} does not, neither for standard nor strong analogy assertions.

\complexity{\subsection{Computational Complexity}
We look at the complexity of concept subsumption w.r.t. $\EL_{\bot}^{\textit{ana}}$ TBoxes for the case when all concepts occurring in the TBox are natural. We concentrate on devising a tight \textsc{coNP} decision procedure; the matching lower bound is inherited from $\EL_\bot^{\bowtie}$~\cite{KR2020interpolation}. For the upper bound, we first show that for any model of the input TBox, one can construct a model in which the number of features, domains and objects in the DL-domain are polynomially bounded in the size of the TBox. The proof requires some care because the number of features and domains of certain concepts occurring in an analogy assertion need to be identical. We then provide a polynomial time guess-and-check procedure.
\begin{theorem}
If all concepts occurring in the input TBox are natural, concept subsumption w.r.t. $\EL_{\bot}^{\textit{ana}}$ TBoxes  is \textsc{coNP}-complete.
\end{theorem}}


\section{Conclusions}
We have proposed a framework for analogy assertions of the form ``$A$ is to $B$ what $C$ is to $D$'' that is suitable for analogical reasoning in description logics. The underlying assumption is that analogy assertions between concept names can be learned from text, and that we can then lift these to obtain analogy assertions between complex DL concepts. We have shown how the resulting semantics allows us to infer concept inclusions using two analogy based inference patterns. This complements other types of plausible inference patterns, such as interpolation and similarity based reasoning. 

There are two important lines for immediate future work. First, we plan to study the computational complexity of reasoning in  $\EL_{\bot}^{\textit{ana}}$. Results from~\cite{KR2020interpolation}  provide a \textsc{coNP} lower bound for concept subsumption w.r.t. $\EL_{\bot}^{\textit{ana}}$ TBoxes. For the upper bound, one key issue would be to relate the number of domains and  features of  concepts occurring in analogy assertions, as well of those of existential restrictions over intra-domain roles of such concepts. One would also need to establish a bound on the number of features and domains. From the practical side we need mechanisms to deal with the noisy nature of the available knowledge about betweenness  and analogy assertions (which typically would be learned from data) and the inconsistencies that may introduce. To this end, we plan to study probabilistic or non-monotonic extensions of our framework.



\ifappendix
\appendix
\section{Proofs}
\subsection*{Proof of Proposition \ref{propIRFBSA1}}
First note that $A_2\sqsubseteq B_2$ is equivalent to $\varphi(B_2)\subseteq \varphi(A_2)$, since $A_2$ and $B_2$ are natural concepts. 
Let $f\in \varphi(B_2)$. We thus need to show that $f\in\varphi(A_2)$. If $f\notin \varphi(B_1)$ then we must have $f\in \varphi(A_2)\setminus \varphi(A_1)$, since $\ap{A_1}{A_2}{B_1}{B_2}$ is satisfied. If $f\in \varphi(B_1)$, then we must have $f\in \varphi(A_1)$ since $A_1 \sqsubseteq B_1$ is satisfied. Since $f\notin \varphi(B_1)\setminus \varphi(B_2)$ we must have $f\notin \varphi(A_1)\setminus \varphi(A_2)$, which means $f\in\varphi(A_2)$.

\subsection*{Proof of Proposition \ref{propDomainTranslationsReversed}}
Let us write $U=\{(s_1,t_1),...,(s_k,t_k)\}$. We show for each $g\in\mathcal{F}$ that $\sigma_{\inv{U}}(\sigma_U(g))=g$. If $g\in \mathcal{F}_{s_i}$, for some $i\in\{1,...,k\}$, then by the definition of domain translation, we have $\sigma_{\inv{U}}(\sigma_U(g))=\sigma_{(t_i,s_i)}(\sigma_{(s_i,t_i)}(g))=g$. If $g\notin \mathcal{F}_{s_1}\cup...\cup \mathcal{F}_{s_k}$, then $\sigma_U(g)=g$. In that case, we also have $g\notin \mathcal{F}_{t_1}\cup...\cup \mathcal{F}_{t_k}$, since $g\in \mathcal{F}_j$ for some $j\in \delta(A_1)\setminus \textit{src}(U)$ and we know from \eqref{eqTargetSourceDisjointness} that $\textit{tgt}(U) \cap (\delta(A_1)\setminus \textit{src}(U))= \emptyset$.
From $g\notin \mathcal{F}_{t_1}\cup...\cup \mathcal{F}_{t_k}$ we find $\sigma_{\inv{X}}(g)=g$ and in particular $\sigma_{\inv{U}}(\sigma_U(g))=\sigma_{\inv{U}}(g)=g$.

\subsection*{Proof of Proposition \ref{propMuComposition}}
We show this proposition in two steps. 
\begin{lemma}\label{lemmaMappingComposition1}
We have
$$
\mu(A,C)\supseteq \{\sigma_{U\oplus V} \,|\, \sigma_{U}\in\mu(A,B), \sigma_{V}\in\mu(B,C)\}
$$
\end{lemma}
\noindent\begin{proof}
Let $\sigma_{U}\in\mu(A,B)$ and $\sigma_{V}\in\mu(B,C)$.
We need to show that $\sigma_{U\oplus V}$ satisfies conditions, \eqref{eqDefSigmaAP}, \eqref{eqNoRedundantMappings} and \eqref{eqTargetSourceDisjointness} w.r.t.\ the concept pair $(A,C)$.

First consider \eqref{eqNoRedundantMappings}. Let $i\in \textit{src}(U\oplus V)$. If $i\in \textit{src}(U)$ then we have $i\in \delta(A)$ since $\sigma_U\in\mu(A,B)$. The only other possibility is that $i\in \textit{src}(V)$ and $i \notin\textit{tgt}(U)$. Since $\sigma_V\in\mu(B,C)$, we have that $i\in \textit{src}(V)$ implies $i\in \delta(B)$. Since $\sigma_U\in\mu(A,B)$ and $i \notin\textit{tgt}(U)$, this furthermore implies $i\in\delta(A)$. In all cases we thus have that \eqref{eqNoRedundantMappings} is satisfied.

Next we show that \eqref{eqTargetSourceDisjointness} is satisfied, i.e.\ that $\textit{tgt}(U\oplus V)\cap (\delta(A)\setminus \textit{src}(U\oplus V))=\emptyset$. Suppose $k\in \textit{tgt}(U\oplus V)$; we show that $k\notin \delta(A)\setminus \textit{src}(U\oplus V)$. We consider two cases:
\begin{itemize}
\item Suppose $k\notin \delta(A)\setminus \textit{src}(U)$. Suppose $k\in \delta(A)\setminus \textit{src}(U\oplus V)$ were to hold. This is only possible if $k\in \textit{src}(U)\setminus \textit{src}(U\oplus V)$. By definition of $\oplus$, this is only possible if $(k,j)\in U$ and $(j,k)\in V$ for some $j$. From $k\in \textit{tgt}(U\oplus V)$ we then find that $(i,k)\in U$ for some $i$ and $k\notin \textit{src}(V)$. However, this implies that $k\in\textit{tgt}(V)\cap (\delta(B) \setminus \textit{src}(V))$, which is a contradiction since $\sigma_V\in\mu(B,C)$. We thus have $k\notin \delta(A)\setminus \textit{src}(U\oplus V)$.
\item Suppose $k\in \delta(A)\setminus \textit{src}(U)$. Then $k\in \delta(B)$. However, since $\sigma_Y\in \mu(B,C)$, we know that $k\notin \delta(B)\setminus \textit{src}(V)$, hence we find $k\in \textit{src}(V)$. In other words, $V$ contains some pair of the form $(k,l)$. Furthermore, from $\textit{tgt}(U)\cap (\delta(A)\setminus \textit{src}(U))=\emptyset$, we know that $k\notin \textit{tgt}(U)$. By construction of $U\oplus V$ we find $(k,l)\in U\oplus V$ and thus $k\in \textit{src}(U\oplus V)$, and in particular $k\notin \delta(A)\setminus \textit{src}(U\oplus V)$.
\end{itemize}

Finally, we show that \eqref{eqDefSigmaAP} is satisfied, i.e.\ that $\varphi(C)=\{\sigma_{U\oplus V}(f)\,|\, f\in \varphi(A)\}$. Let $f\in\varphi(A)$.
\begin{itemize}
\item Suppose $(i,j)\in U$, $(j,k)\in V$ and $i\neq k$. Since $\sigma_U\in\mu(A,B)$ and $\sigma_V\in\mu(B,C)$, we know that $\sigma_{(i,j)}(f)\in \varphi(B)$ and $\sigma_{(j,k)}(\sigma_{(i,j)}(f))\in \varphi(C)$. Since $\sigma_{(j,k)}\circ \sigma_{(i,j)}=\sigma_{(i,k)}$ we find $\sigma_{(i,k)}(f)\in \varphi(C)$, and in particular $\sigma_{U\oplus V}(f)\in\varphi(C)$.
\item Suppose $(i,j)\in X$ and $j\notin \textit{src}(V)$. We then have $\sigma_{(i,j)}(f)\in \varphi(B)$ as well as $\sigma_{(i,j)}(f)\in \varphi(C)$, and in particular $\sigma_{U\oplus V}(f)\in\varphi(C)$.
\item Suppose $(j,k)\in Y$ and $j\notin \textit{tgt}(U)$. This implies that $j\notin \textit{src}(U)$. We thus have $f\in\varphi(B)$, $\sigma_{(j,k)}(f)\in \varphi(C)$ and in particular $\sigma_{U\oplus V}(f)\in \varphi(C)$.
\end{itemize}
This already shows that $\varphi(C)\supseteq\{\sigma_{U\oplus V}(f)\,|\, f\in \varphi(A)\}$. Conversely, suppose $f\in \varphi(C)$. Then we know there must be some $g\in \varphi(B)$ such that $\sigma_V(g)=f$ and some $h\in \varphi(A)$ such that $\sigma_U(h)=g$. Suppose $h\in\mathcal{F}_i$, $g\in\mathcal{F}_j$ and $f\in\mathcal{F}_k$.
\begin{itemize}
\item If $h\neq g$ and $g\neq f$, it must be the case that $(i,j)\in U$ and $(j,k)\in V$. If $i\neq k$ we have $(i,k)\in U\oplus V$ and thus $\sigma_{U\oplus V}(h)=f$. If $i=k$, we have $i\notin \textit{src}(U\oplus V)$ and thus again $\sigma_{U\oplus V}(h)=f$.
\item If $h\neq g$ and $g=f$, we have $j=k$, $j\notin \textit{src}(V)$ and thus $(i,j)\in U\oplus V$, meaning $\sigma_{U\oplus V}(h)=\sigma_U(h)=g=f$.
\item If $h=g$ and $g\neq f$, we have $i=j$, $j\notin \textit{tgt}(U)$ and thus $(j,k)\in U\oplus V$, meaning $\sigma_{U\oplus V}(h)=\sigma_{U\oplus V}(g)=\sigma_V(g)=f$.
\end{itemize}
This shows $\varphi(C)\subseteq\{\sigma_{U\oplus V}(f)\,|\, f\in \varphi(A)\}$
\end{proof}
\noindent To complete the proof of Proposition \ref{propMuComposition}, we now also show the following result
\begin{lemma}\label{lemmaMuMappingComposition}
If $\mu(A,B)\neq \emptyset$ and $\mu(B,C)\neq \emptyset$ then it holds that:
$$
\mu(A,C) \subseteq \{\sigma_{U\oplus V} \,|\, \sigma_U\in\mu(A,B), \sigma_V\in\mu(B,C)\}
$$
\end{lemma}
\noindent \begin{proof}
Let $\sigma_Z\in\mu(A,C)$. We show that there are mappings $\sigma_{U^*}\in\mu(A,B)$ and $\sigma_{V^*}\in\mu(B,C)$ such that $Z=U^*\oplus V^*$.

Let $\sigma_U$ and $\sigma_V$ be arbitrary elements from $\mu(A,B)$ and $\mu(B,C)$ respectively. If $Z=U\oplus V$ then we can simply choose $U^*=U$ and $V^*=V$. Now suppose there is some feature $f\in\mathcal{F}_i$ such that $\sigma_{U\oplus V}(f)\neq \sigma_Z(f)$. Let us write $g = \sigma_{U\oplus V}(f)$ and $g' = \sigma_{Z}(f)$. Furthermore, let $f'$ and $f''$ be the features from $\varphi(A)$ such that $\sigma_{U\oplus V}(f')=g'$ and $\sigma_Z(f'')=f$. Let us write $i,i',i''$ for the domains of $f,f',f''$ and $j,j'$ for the domains of $g,g'$ respectively. Note that we then have
\begin{align*}
\varphi(A)\cap\mathcal{F}_i &= \{\sigma_{U\oplus V}^{-1}(x) \,|\, x\in \varphi(C)\cap \mathcal{F}_j\}\\
&=\{\sigma_{Z}^{-1}(x) \,|\, x\in \varphi(C)\cap \mathcal{F}_{j'}\}
\end{align*}
from which we find
\begin{align*}
\varphi(C)\cap \mathcal{F}_{j'} &= \{\sigma_{Z}(\sigma_{U\oplus V}^{-1}(x)) \,|\, x\in \varphi(C)\cap \mathcal{F}_j\}\\
\varphi(C)\cap \mathcal{F}_{j} &= \{\sigma_{U\oplus V}(\sigma_{Z}^{-1}(x)) \,|\, x\in \varphi(C)\cap \mathcal{F}_{j'}\}
\end{align*}
and in particular thus also:
\begin{align}
\varphi(C)\cap \mathcal{F}_{j'} &= \{\sigma_{(j,j')}(x) \,|\, x\in \varphi(C)\cap \mathcal{F}_j\} \label{eqlemmaMuMappingCompositionA}\\
\varphi(C)\cap \mathcal{F}_{j} &= \{\sigma_{(j',j)}(x) \,|\, x\in \varphi(C)\cap \mathcal{F}_{j'}\}\label{eqlemmaMuMappingCompositionB}
\end{align}
We show that we can always find $U'$ and $V'$ such that $\sigma_{U'}\in\mu(A,B)$, $\sigma_{V'}\in\mu(B,C)$,  $\sigma_{U'\oplus V'}(f)= \sigma_Z(f)$, and such that for each feature $f'$ for which we had $\sigma_{U\oplus V}(f')= \sigma_Z(f')$, we have  $\sigma_{U'\oplus V'}(f')= \sigma_Z(f')$. Thus, given that the set of features $\mathcal{F}$ is finite, by repeating the same process, we will end up with mappings $\sigma_{U^*}$ and  $\sigma_{V^*}$ such that $\sigma_{U^*\oplus V^*}=\sigma_Z$.
\begin{itemize}
\item Suppose $U$ contains a pair of the form $(i,l)$ and $V$ contains pair of the form $(l,j)$.
\begin{itemize}
\item Suppose $V$ also contains a pair of the form $(l',j')$. Let $U'=U$ and $V'=(V\setminus\{(l,j),(l',j')\})\cup \{(l,j'),(l',j)\}$. From \eqref{eqlemmaMuMappingCompositionA}--\eqref{eqlemmaMuMappingCompositionB}, together with the transitivity properties of the mappings $\sigma_{(i,j)}$ (i.e.\ Condition 3 from Definition \ref{defDomainInterpretation}), it follows that 
\begin{align*}
\varphi(C)\cap \mathcal{F}_{j'} &= \{\sigma_{(l,j')}(x) \,|\, x\in \varphi(B)\cap \mathcal{F}_l\}\\
\varphi(C)\cap \mathcal{F}_{j} &= \{\sigma_{(l,j)}(x) \,|\, x\in \varphi(B)\cap \mathcal{F}_{l'}\}
\end{align*}
It follows that $\sigma_{V'}\in\mu(B,C)$ while $U'$ and $V'$ satisfy the required conditions.
\item Suppose $V$ does not contain any pair of the form $(l',j')$ but $U$ contains a pair of the form $(l',j')$, then we define $U'=(U\setminus\{(l',j')\})\cup \{(l',j)\}$ and $V'=(V\setminus\{(l,j)\})\cup \{(l,j')\}$. Similarly as in the previous case, we find that $U'$ and $V'$ satisfy the requirements.
\item If $U$ does not contain any pair of the form $(l',j')$ either, then it must be the case that $\sigma_{U\oplus V}(f')=f'$ and $i'=j'$. In this case, we can choose $U'=U$ and $V'=(V\setminus\{(l,j)\})\cup \{(l,j'),(i',j)\}$.
\end{itemize}
\item Suppose $U$ contains a pair of the form $(i,j)$.
\begin{itemize}
\item If $V$ contains a pair of the form $(l',j')$, then we can choose $U'=(U\setminus \{(i,j)\}) \cup \{(i,j')\}$ and $V'=(V\setminus \{(l',j')\}) \cup \{(l',j)\}$.
\item If $V$ does not contain a pair of the form $(l',j')$, but $U$ contains such a pair then we can choose $U'=(U\setminus \{(i,j),(l',j')\}) \cup \{(i,j'),(l',j)\}$ and $V'=V$.
\item If $U$ does not contain a pair of the form $(l',j')$ either, then $i'=j'$ and we can choose $U'=(U\setminus \{(i,j)\}) \cup \{(i,j')\}$ and $V'=V\cup \{(i',j)\}$.
\end{itemize}
\item Suppose $V$ contains a pair of the form $(i,j)$.
\begin{itemize}
\item Suppose $V$ also contains a pair of the form $(l',j')$. Then we can choose $U'=U$ and $V'=(V\setminus\{(i,j),(l',j')\})\cup \{(i,j'),(l',j)\}$.
\item Suppose $V$ does not contain a pair of the form $(l',j')$ but $U$ contains such a pair. Then we can choose $U'=(U\setminus\{(l',j')\})\cup \{(l',j)\}$ and $V'=(V\setminus\{(i,j)\})\cup \{(i,j')\}$.
\item Suppose $U$ does not contain a pair of the form $(l',j')$ either. Then $i'=l'$ and we can choose $U'=U$ and $V'=(V\setminus\{(i,j)\})\cup \{(i,j'),(i',j)\}$.
\end{itemize}
\end{itemize}
We thus find that suitable sets $U'$ and $V'$ can be found in all cases.
\end{proof}

\subsection*{Proof of Proposition \ref{propStrictUniqueDomainMapping}}
We first show the following lemmas.

\begin{lemma}\label{lemmaStrictEmptyConcept}
Let $\Imf= ( \Imc, [\mathcal{F}_1,...\mathcal{F}_k], \mathcal{X}, \pi, \anaEquiv, \mathcal{S})$ be a domain-constrained interpretation.
If $i,j\in \delta(A)$ such that $(i,j)\in \anaEquiv$ and $i\neq j$, it holds that $A^{\Imc}=\emptyset$.
\end{lemma}
\noindent \begin{proof}
This follows immediately from Conditions 1 and 5 of Definition \ref{defDomainInterpretation}.
\end{proof}

\begin{lemma}\label{lemmaStricInterpretationMuMinimal}
Let $\Imf= ( \Imc, [\mathcal{F}_1,...\mathcal{F}_k], \mathcal{X}, \pi, \anaEquiv, \mathcal{S})$ be a domain-constrained interpretation. If $\mu(A,B)\neq \emptyset$ then there is some $\sigma_u\in\mu(A,B)$ such that for every $\sigma_V\in \mu(A,B)$ it holds that $U\subseteq V$. 
\end{lemma}
\noindent \begin{proof}
Suppose that $\sigma_U,\sigma_V\in\mu(A,B)$ such that $U$ and $V$ are minimal, i.e.\ for any $U'\subset U$ or $V'\subset V$ we have $\sigma_{U'}\notin \mu(A,B)$ and $\sigma_{V'}\notin \mu(A,B)$. Now suppose $U\neq V$, i.e.\ suppose that there is some pair $(i_1,i_2)\in U\setminus V$ and some pair $(j_1,j_2)\in V\setminus U$. 

First assume $i_1=j_1$. Since $U$ and $V$ were assumed to be minimal, we have $i_1,j_1\in \delta(A)$ and thus we must also have $i_2,j_2\in\delta(B)$. However, that means $\varphi(B)$ contains some $f\in \mathcal{F}_{i_2}$ and some $g\in \mathcal{F}_{j_2}$, while $(i_2,j_2)\in \anaEquiv$. We thus find from Lemma \ref{lemmaStrictEmptyConcept} that $B^{\Imc}=\emptyset$, or in other words that $\varphi(B)=\mathcal{F}$. This also entails that $\varphi(A)=\mathcal{F}$. The only way to choose $\sigma_U$ and $\sigma_V$ such that $U$ and $V$ are minimal is thus to choose $U=V=\emptyset$.

Now assume there is no pair $(j_1,j_2)\in V\setminus U$ such that $i_1=j_1$. Since $U$ was assumed to be minimal, we must have some $f\in \mathcal{F}_{i_1}\cap\varphi(A)$. Since, by assumption, we have no pair of the form $(i_1,j)$ in $V$, we must also have $f\in \varphi(B)$. We thus find that $f$ and $\sigma_U(f)$ both belong to $\varphi(B)$, but $f\in \mathcal{F}_{i_1}$ and $\sigma_U(f)\in \mathcal{F}_{i_2}$. Since $U$ was assumed to be minimal, we also have $i_1\neq i_2$, hence it again follows that $B^{\Imc}=\emptyset$, and thus $\varphi(A)=\varphi(B)=\mathcal{F}$ and $U=V=\emptyset$.
\end{proof}
From Lemma \ref{lemmaStricInterpretationMuMinimal}, we know that $\mu(A,B)$ contains some element $\sigma_U$ such that $U\subseteq V$ for each $\sigma_V\in \mu(A,B)$. Now suppose $U\neq V$ and let $(i,j)\in V\setminus U$. Note that by \eqref{eqNoRedundantMappings} we have that $i\in \delta(A)$. We can thus only have $\sigma_U(\varphi(A))= \sigma_V(\varphi(A))$ if $U$ contains a pair of the form $(i,k)$. That means that $i,k\in \delta(B)$ with $i\neq k$, and thus $A^{\Imc}=B^{\Imc}=\emptyset$ (using Lemma \ref{lemmaStrictEmptyConcept}).

\subsection*{Proof of Proposition \ref{propAnaReversible}}
We first show the following lemma.
\begin{lemma}\label{lemmaDefSigmaDisjointnessInv}
Suppose that $\sigma_X\in \mu(A_1,A_2)$. Then $\sigma_{\inv{X}} \in \mu(A_2,A_1)$.
\end{lemma}
\noindent\begin{proof}
We first show that 
$$
\textit{tgt}(\inv{X})\cap (\delta(A_2)\setminus \textit{src}(\inv{X}) )=\emptyset
$$
which is equivalent to:
$$
\textit{src}(X)\cap (\delta(A_2)\setminus \textit{tgt}(X) )=\emptyset
$$
Let $i\in \delta(A_2)\setminus \textit{tgt}(X)$. Then there must be some $f\in \varphi(A_2)$ such that $f\in \mathcal{F}_i$. But since $i\notin \textit{tgt}(X)$ this is only possible if $f\in \varphi(A_1)$ and $i\notin \textit{src}(X)$. We thus have $\textit{src}(X)\cap (\delta(A_2)\setminus \textit{tgt}(X) )=\emptyset$. 
We now show that $\varphi(A_1) = \{ \sigma_{\inv{X}}(f) \,|\, f\in \varphi(A_2)\}$. Since $\sigma_X \in \mu(A_1,A_2)$, this means that we need to show: 
$$
\varphi(A_1) = \{ \sigma_{\inv{X}}(\sigma_X(g)) \,|\, g\in \varphi(A_1)\}
$$
which follows immediately from Proposition \ref{propDomainTranslationsReversed}. Finally, it is also clear that \eqref{eqNoRedundantMappings} is satisfied for $\sigma_{\inv{X}}$ if this condition is satisfied for $\sigma_{X}$
\end{proof}
Since $\inv{(\inv{U})}=U$, we have the following corollary.
\begin{corollary}\label{corInvertibleMapping}
It holds that:
\begin{align*}
\mu(A_2,A_1) = \{\sigma_{\inv{U}} \,|\, \sigma_U\in\mu(A_1,A_2)\}
\end{align*}
\end{corollary}
\noindent The main result directly follows from this corollary.

\subsection*{Proof of Proposition \ref{propTransitivity}}
We show both transitivity properties separately.
\begin{lemma}
It holds that:
$$
\{\ana{A_1}{A_2}{B_1}{B_2},\ana{B_1}{B_2}{C_1}{C_2}\} \models \ana{A_1}{A_2}{C_1}{C_2}
$$
\end{lemma}
\noindent \begin{proof}
Assume that $\ana{A_1}{A_2}{B_1}{B_2}$ and $\ana{B_1}{B_2}{C_1}{C_2}$ are satisfied in some domain-constrained interpretation $\Imf$. We then have that there is some $\sigma_U\in \mu(A_1,A_2)\cap\mu(B_1,B_2)$ and some  $\sigma_V\in \mu(B_1,B_2)\cap\mu(C_1,C_2)$. 
By Proposition \ref{propStrictUniqueDomainMapping}, we have $\sigma_U=\sigma_V$ and thus we find that $\mu(A_1,A_2)\cap\mu(C_1,C_2)\neq \emptyset$.
\end{proof}

\begin{lemma}
It holds that:
$$
\{\ana{A_1}{A_2}{B_1}{B_2},\ana{A_2}{A_3}{B_2}{B_3}\} \models \ana{A_1}{A_3}{B_1}{B_3}
$$
\end{lemma}
\noindent \begin{proof}
Assume that $\ana{A_1}{A_2}{B_1}{B_2}$ and $\ana{A_2}{A_3}{B_2}{B_3}$ are satisfied in some domain-constrained interpretation $\Imf$. Then there exists some $\sigma_U\in \mu(A_1,A_2)\cap\mu(B_1,B_2)$ and some  $\sigma_V\in \mu(A_2,A_3)\cap\mu(B_2,B_3)$. By Proposition \ref{propMuComposition} we then have that $\sigma_{U\oplus V}\in\mu(A_1,A_3) \cap \mu(B_1,B_3)$, and hence that $\ana{A_1}{A_3}{B_1}{B_3}$ is satisfied.
\end{proof}

\subsection*{Proof of Proposition \ref{propCTransitivity}}
Assume that $\ana{A_1}{B_1}{B_2}{A_2}$ and $\ana{A_1}{C_1}{C_2}{A_2}$ are satisfied in some domain-constrained interpretation $\Imf$. There exist some $\sigma_U\in \mu(A_1,B_1)\cap \mu(B_2,A_2)$ and some $\sigma_V\in \mu(A_1,C_1)\cap \mu(C_2,A_2)$.

First assume that $A_1^{\Imc}=\emptyset$. Then clearly we also have $B_1^{\Imc}=C_1^{\Imc}=\emptyset$, since $\varphi(A_1)=\mathcal{F}$ and $|\varphi(A_1)|=|\varphi(B_1)|=|\varphi(C_1)|$. If moreover $A_2^{\Imc}=\emptyset$, then we similarly also find $B_2^{\Imc}=C_2^{\Imc}=\emptyset$ and the result is trivially satisfied. If $A_2^{\Imc}\neq\emptyset$, then the only possibility is that $U=V=\emptyset$. Indeed, if e.g.\ $(i,k)\in V$ then $V$ would need to contain some element of the form $(j,i)$ with $(i,j)\in \anaEquiv$, which would entail $C_2^{\Imc}=\emptyset$ since $\textit{src}(V)\subseteq \delta(C_2)$ and $i,j\in\delta(C_2)$ implies $C_2^{\Imc}=0$ by Lemma \ref{lemmaStrictEmptyConcept}. Thus we have $A_2^{\Imc}=B_2^{\Imc}=C_2^{\Imc}$, meaning that the result is satisfied. 
The case where $A_2^{\Imc}=\emptyset$ and $A_1^{\Imc}\neq\emptyset$ is analogous.

We now show that the result holds for the case where $A_1^{\Imc}\neq\emptyset$ and $A_2^{\Imc}\neq\emptyset$.
 From Corollary \ref{corInvertibleMapping} and Proposition \ref{propMuComposition} we know that then $\mu_{\inv{U}\oplus V}\in \mu(B_1,C_1)$ and $\mu_{V\oplus \inv{U}}\in \mu(C_2,B_2)$. To complete the proof, we show that $\mu_{\inv{U}\oplus V}=\mu_{V\oplus \inv{U}}=\mu_{\inv{U}\cup V}$, from which it follows in particular that $\mu_{\inv{U}\oplus V}\in \mu(B_1,C_1)\cap \mu(C_2,B_2)$, meaning that $\ana{A_1}{A_3}{B_1}{B_3}$ is satisfied.

Suppose that there are elements $i,j,k$ such that $(i,j)\in \inv{U}$ and $(j,k)\in V$. Then we have $i\in \delta(A_2)$ since $\textit{src}(\inv{U})\subseteq \delta(A_2)$. However, we also have $k\in\delta(A_2)$ since $\textit{src}(V)\subseteq \delta(C_2)$ and $\varphi(A_2)=\sigma_V(\varphi(C_2))$. Since $(i,k)\in\anaEquiv$ and we assumed $A_2^{\Imc}\neq\emptyset$, it follows that $i=k$. Since this is the case for all $i,j,k$ such that $(i,j)\in \inv{U}$ and $(j,k)\in V$, it follows from the definition of $\oplus$ that $\inv{U}\oplus V=\inv{U}\cup V$. In the same way we find that $V\oplus\inv{U}=\inv{U}\cup V$.

\subsection*{Proof of Proposition \ref{propLiftingConjunction}}
We first show the following lemma. 
\begin{lemma}\label{lemmaProofPropLiftingConjunction}
Let $\Imf=( \Imc,  [\mathcal{F}_1,...,\mathcal{F}_k], \mathcal{X}, \pi, \anaEquiv, \mathcal{S})$ be a domain-constrained interpretation satisfying $(A_1 {\sqcap} B_1)^{\Imc}\neq \emptyset$ and $(A_2 {\sqcap} B_2)^{\Imc}\neq \emptyset$. Let $\sigma_U\in\mu(A_1,B_1)$ and $\sigma_V\in\mu(A_2,B_2)$. It holds that $\sigma_{U\cup V} \in \mu(A_1 \sqcap B_1,A_2 \sqcap B_2)$.  
\end{lemma}
\noindent\begin{proof}
We first show that $\sigma_{U\cup V}$ is a valid domain translation:
\begin{itemize}
\item All pairs in $U\cup V$ should have a different source domain, i.e.\ we must have that for all $(i,j)\in U$ and $(i,k)\in V$ it holds that $j=k$. To see why this is satisfied, note that $(i,j)\in U$ and $(i,k)\in V$ imply that $j\in\delta(A_2)$ and $k\in\delta(B_2)$. From Lemma \ref{lemmaStrictEmptyConcept}, we know that $j\neq k$ would imply $(A_2\sqcap B_2)^{\Imc}=\emptyset$.
\item All pairs in $U\cup V$ should have a different target domain, i.e.\ for all $(i,k)\in U$ and $(j,k)\in V$ we must have that $i=j$. To see why this is satisfied, note that $(i,k)\in U$ and $(j,k)\in V$ mean $i\in \delta(A_1)$ and $j\in \delta(B_1)$. We thus find that $i\neq j$ would imply $(A_1\sqcap B_1)^{\Imc}=\emptyset$.
\end{itemize}
Next we show that $\textit{tgt}(U\cup V)\cap (\delta(A_1\sqcap B_1) \setminus \textit{src}(U\cup V)) = \emptyset$. Suppose that there were some $i\in \textit{tgt}(U\cup V)\cap (\delta(A_1\sqcap B_1) \setminus \textit{src}(U\cup V))$. In other words, suppose we had $i\in \textit{tgt}(U)\cup \textit{tgt}(V)$, $i\in \delta(A_1)\cup \delta(B_1)$, $i\notin \textit{src}(U)$ and $i\notin \textit{src}(V)$. Suppose in particular that $i\in \textit{tgt}(U)$; the case where $i\in \textit{tgt}(V)$ is entirely analogous. Since $\sigma_U\in\mu(A_1,A_2)$ and $i\notin\textit{src}(U)$ we know that $i\notin \delta(A_1)$, and thus in particular that $i\in \delta(B_1)$. From $i\in \textit{tgt}(U)$,  we know that there must be some $(j,i)\in U$ such that $j\in\delta(A_1)$ and $(j,i)\in\anaEquiv$. Together with $i\in\delta(B_1)$ we find $i,j\in \delta(A_1\sqcap B_1)$, which would imply $(A_1\sqcap B_1)^{\Imc}=\emptyset$, using Lemma \ref{lemmaStrictEmptyConcept}.

Next we show that for $f\in \varphi(A_1)$ it must be the case that $\sigma_{U\cup V}(f)=\sigma_U(f)$. Assume $f\in\mathcal{F}_i$. If $i\in\textit{src}(U)$ then there is some element $(i,j)\in U$ and $\sigma_{U\cup V}(f)=\sigma_U(f) = \sigma_{(i,j)}(f)$. If $i\notin \textit{src}(U\cup V)$ then we trivially have $\sigma_{U\cup V}(f)=\sigma_U(f)=f$. Finally, assume that $i\in \textit{src}(V)\setminus \textit{src}(U)$, then $V$ contains some element $(i,j)$ from $\anaEquiv$ such that $i\in\delta(A_2)$ and $j\in \delta(B_2)$, but this would imply $(A_2\sqcap B_2)^{\Imc}=\emptyset$ using Lemma \ref{lemmaStrictEmptyConcept}. Similarly, we also find that for $f\in \varphi(B_1)$ it must be the case that $\sigma_{U\cup V}(f)=\sigma_V(f)$. It follows that $\varphi(A_2\sqcap B_2) = \{\sigma_{U\cup V}(f) \,|\, f\in \varphi(A_1\sqcap B_1)\}$. Indeed:
\begin{align*}
& \{\sigma_{U\cup V}(f) \,|\, f\in \varphi(A_1\sqcap B_1)\}\\
 &=\{\sigma_{U\cup V}(f) \,|\, f\in \varphi(A_1)\} \cup \{\sigma_{U\cup V}(f) \,|\, f\in \varphi(B_1)\}\\
 &=\{\sigma_{U}(f) \,|\, f\in \varphi(A_1)\} \cup \{\sigma_{V}(f) \,|\, f\in \varphi(B_1)\}\\
 &=\varphi(A_2) \cup \varphi(B_2)\\
 &=\varphi(A_1\sqcap B_2)
\end{align*}
\end{proof}
Now we return to the main result. From $\ana{A_1}{A_2}{A_3}{A_4}$ and $\ana{B_1}{B_2}{B_3}{B_4}$ we know that there exists some
 $\sigma_U \in \mu(A_1,A_2)\cap \mu(A_3,A_4)$ and $\sigma_V \in \mu(B_1,B_2)\cap \mu(B_3,B_4)$. From Lemma \ref{lemmaProofPropLiftingConjunction} we then find $\sigma_{U\cup V} \in \mu(A_1\sqcap B_1,A_2\sqcap B_2)\cap \mu(A_3\sqcap B_2,A_4\sqcap B_4)$, which means that $\ana{(A_1 \sqcap B_1)}{(A_2 \sqcap B_2)}{(A_3 \sqcap B_3)}{(A_4 \sqcap B_4)}$ is satisfied.

\subsection*{Proof of Proposition \ref{propLiftingExistential}}


\begin{lemma}\label{lemmaIntraRoleMuInclusion}
Let $\Imf= ( \Imc,  [\mathcal{F}_1,...,\mathcal{F}_k], \mathcal{X}, \pi, \anaEquiv, \mathcal{S})$ be a domain constrained interpretation and let $r$ be an intra-domain role. It holds that $\mu(A,B)\subseteq\mu(\exists r.A,\exists r.B)$.
\end{lemma}
\noindent \begin{proof}
Let $\sigma_U\in\mu(A,B)$. We first show that $\varphi(\exists r.B)=\sigma_U(\varphi(\exists r.A))$. We find:
\begin{align*}
\sigma_U(\varphi(\exists r.A)) 
&=\sigma_U(\kappa_r(\varphi(A)))\\
&=\kappa_r(\sigma_U(\varphi(A)))\\
&=\kappa_r(\varphi(B))\\
&=\varphi(\exists r.B)
\end{align*}
Finally, the fact that $\textit{src}(U)\subseteq \delta(\exists r.A)$ and $\textit{tgt(U)} \cap (\delta(\exists r.A) \setminus \textit{src}(U))= \emptyset$ are satisfied, follows from the fact that $\sigma_U\in\mu(A,B)$ and the fact that $\delta(\exists r.A)=\delta(A)$, where the latter follows immediately from the definition of intra-domain relation.
\end{proof}

\noindent We now show that \eqref{eqLiftingExistential1} and \eqref{eqLiftingExistential2} hold. Since $\ana{A}{B}{C}{D}$ is satisfied, there must exist some $\sigma_U\in \mu(A,B)\cap \mu(C,D)$. From Lemma \ref{lemmaIntraRoleMuInclusion}, we find that $\sigma_U\in \mu(\exists r.A,\exists r.B)\cap \mu(\exists r.C,\exists r.D)$, and thus that \eqref{eqLiftingExistential1} is satisfied. Similarly, we also find that $\sigma_U\in \mu( A,B)\cap \mu(\exists r.C,\exists r.D)$, and thus that \eqref{eqLiftingExistential2} is satisfied.

\subsection*{Proof of Proposition \ref{propSubsumptionTranslation1}}
Assume that $\Imf$ satisfies $\ana{A_1}{B_1}{A_2}{B_2}$ and $A_1 \sqsubseteq A_2$. We need to show that $\Imf$ satisfies $B_1 \sqsubseteq B_2$, or equivalently, that $\varphi(B_2)\subseteq \varphi(B_1)$. Let $f\in \varphi(B_2)$. Since $\Imf$ satisfies $\ana{A_1}{B_1}{A_2}{B_2}$ there exists some $\sigma_U\in\mu(A_1,B_2)\cap\mu(A_2,B_2)$. Hence there must be some $g\in\varphi(A_2)$ such that $\sigma_U(g)=f$. Since $A_1\sqsubseteq A_2$ is satisfied, we have $g\in \varphi(A_1)$, which in turn implies that $\sigma_U(g)\in\varphi(B_1)$, i.e.\ $f\in\varphi(A_1)$.

\subsection*{Proof of Proposition \ref{propSubsumptionExtrapolation1}}
If $A_4^{\Imc}= \emptyset$ then the conclusion holds trivially. Moreover, if $A_3^{\Imc}= \emptyset$ then we have $\varphi(A_3)=\mathcal{F}$, hence from $\ana{A_1}{A_2}{A_3}{A_4}$ we find $\varphi(A_4)=\mathcal{F}$ and thus $A_4^{\Imc}= \emptyset$, hence again the conclusion holds trivially. From $A_1^{\Imc}\neq \emptyset$ we similarly find $A_2^{\Imc}\neq \emptyset$.  
If $B_4^{\Imc}= \emptyset$ then we also have $B_3^{\Imc}= \emptyset$, which would imply $A_3^{\Imc}= \emptyset$ and thus also $A_4^{\Imc}= \emptyset$, meaning that the conclusion is again trivially satisfied. The same holds whenever $B_3^{\Imc}= \emptyset$. Finally, since $A_1^{\Imc}\neq \emptyset$, it must be the case that $A_2^{\Imc}\neq \emptyset$, $B_1^{\Imc}\neq \emptyset$ and $B_2^{\Imc}\neq \emptyset$.
In the following, we can thus assume w.l.o.g.\ that $A_i^{\Imc}\neq \emptyset$ and $B_i^{\Imc}\neq \emptyset$ for $i\in\{1,2,3,4\}$

Suppose that $\Imf$ satisfies the stated condition. Let $\sigma_U\in\mu(A_1,A_2)\cap\mu(A_3,A_4)$, $\sigma_{U'}\in\mu(A_1,A_3)\cap\mu(A_2,A_4)$, $\sigma_V\in\mu(B_1,B_2)\cap\mu(B_3,B_4)$ and $\sigma_{V'}\in\mu(B_1,B_3)\cap\mu(B_2,B_4)$. We need to show that $\Imc$ satisfies $A_4 \sqsubseteq B_4$, which is equivalent to $\varphi(B_4)\subseteq \varphi(A_4)$. Let $f\in\varphi(B_4)$, with $f\in\mathcal{F}_i$. Then there must be some $g\in \varphi(A_3)$ such that $\sigma_Y(g)=f$. 
\begin{itemize}
\item If $f=g$, then it must be the case that either $i\in\delta(B_1)\cap\delta(B_2)$ or that $i\notin\delta(B_1)\cup \delta(B_2)$.
\begin{itemize}
\item Suppose $i\in\delta(B_1)\cap\delta(B_2)$, then there some $h\in \varphi(B_1)\cap\varphi(B_2)\cap \mathcal{F}_i$. Since $A_1 \sqsubseteq B_1$ and $A_2  \sqsubseteq B_2$ are satisfied, that means $h\in \varphi(A_1)\cap\varphi(A_2)$, and in particular $i\in \delta(A_1)\cap\delta(A_2)$, which implies $i\notin \textit{src}(U)$. From $f\in \varphi(B_3)$, we find $f\in \varphi(A_3)$, since $A_3 \sqsubseteq B_3$ is satisfied, which in turn entails $f\in\varphi(A_4)$ since $i\notin \textit{src}(U)$ and $\sigma_U\in\mu(A_3,A_4)$.
\item Suppose $i\notin\delta(B_1)\cup \delta(B_2)$. If $i\notin \textit{src}(U)$, we find $f\in\varphi(A_4)$ as in the previous case. Now suppose $i\in \textit{src}(U)$. Since $\sigma_{U'}\in \mu(B_1,B_3)$ there must be some $(j,i)\in U'$ such that $j\in \delta(B_1)$. Note that $i\notin\delta(B_1)$ hence $i\neq j$. Since $A_1\sqsubseteq B_1$ is satisfied, we then also have $j\in \delta(A_1)$. However, from $i\in\textit{src}(U)$ it follows that $i\in\delta(A_1)$. Using Lemma \ref{lemmaStrictEmptyConcept}, from $i,j\in \delta(A_1)$ and $i\neq j$ we find $A_1^{\Imc}=\emptyset$, which contradicts our assumption that $A_1$ was non-empty. Hence the case where $i\in \textit{src}(U)$ cannot occur.
\end{itemize}
\item Now suppose $f\neq g$, and let $g\in \mathcal{F}_j$. Then we have $(j,i)\in U$, with $i\neq j$, and thus $j\in\delta(B_1)$ and $i\in\delta(B_2)$. Since $A_1 \sqsubseteq B_1$, $A_2  \sqsubseteq B_2$ and $A_3 \sqsubseteq B_3$ are satisfied, we also have $j\in \delta(A_1)\cap\delta(A_3)$ and $i\in \delta(A_2)$. From $i\neq j$ and Lemma \ref{lemmaStrictEmptyConcept}, it follows that $i\notin\delta(A_1)$ and $j\notin\delta(A_2)$. Moreover, by Lemma \ref{lemmaStrictEmptyConcept} and the non-emptiness of $A_1$, there cannot be any $l\in \delta(A_1)$ such that $l\neq j$ and $(l,i)\in\anaEquiv$, as this would also imply $(l,j)\in\anaEquiv$. We thus have that $(j,i)\in U$. Given that $g\in \varphi(B_3)$ and the fact that $A_3\sqsubseteq B_3$ is satisfied, we find $g\in \varphi(A_3)$. Since $(j,i)\in U$ and $\sigma_{U}\in\mu(A_3,A_4)$, it follows that $f\in \varphi(A_4)$.
\end{itemize}

\section{Counterexamples}
\subsection*{Counterexamples for Section \ref{secDesiderata}}
The following example shows that rule extrapolation is not valid for analogical proportions.
\begin{example}
Consider a feature-enriched interpretation $\Imf$ s.t.:
\begin{align*}
\varphi(A_1)&=\varphi(A_2)=\varphi(C_2)=\varphi(C_4)=\{f\}\\
\varphi(A_3)&=\varphi(A_4)=\varphi(C_1)=\varphi(C_3)=\emptyset
\end{align*}
Then $\ap{A_1}{A_2}{A_3}{A_4}$, $\ap{C_1}{C_2}{C_3}{C_4}$, $A_1\sqsubseteq C_1$, $A_2\sqsubseteq C_2$ and $A_3\sqsubseteq C_3$ are satisfied in $\Imf$, whereas $A_4\sqsubseteq C_4$ is not.
\end{example}

\noindent The following example shows that from $\ap{A_1}{B_1}{C_1}{D_1}$ and $\ap{A_2}{B_2}{C_2}{D_2}$, in general we cannot infer $\ap{(A_1\sqcap A_2)}{(B_1\sqcap B_2)}{(C_1\sqcap C_2)}{(D_1\sqcap D_2)}$.
\begin{example}
Consider a feature-enriched interpretation $\Imf$ s.t.:
\begin{align*}
\varphi(A_1)=\varphi(A_2)=\varphi(B_1)=\varphi(C_2)=\emptyset\\ \varphi(C_1)=\varphi(D_1)=\varphi(B_2)=\varphi(D_2)=\{f\}
\end{align*}
Then $\ap{A_1}{B_1}{C_1}{D_1}$ and $\ap{A_2}{B_2}{C_2}{D_2}$ are satisfied in $\Imf$, while $\ap{(A_1\sqcap A_2)}{(B_1\sqcap B_2)}{(C_1\sqcap C_2)}{(D_1\sqcap D_2)}$ is not, since
\begin{align*}
\varphi(B_1\sqcap B_2)\setminus \varphi(A_1\sqcap A_2)&=\{f\}\\
\varphi(D_1\sqcap D_2)\setminus \varphi(C_1\sqcap C_2)&=\emptyset
\end{align*}
\end{example}

\subsection*{Counterexamples for Section \ref{secPropertiesAnaAssertions}}
We show in the following counterexample that the condition $\ana{B_1}{B_3}{B_2}{B_4}$ in Proposition \ref{propSubsumptionExtrapolation1} is indeed required.
\begin{example}
Let $\Imf= ( \Imc,  [\mathcal{F}_1,\mathcal{F}_2], \mathcal{X}, \pi, \anaEquiv, \mathcal{S})$ be a domain constrained interpretation, where $\mathcal{X}=\{\mathcal{F}\}$ and the sets of features $\mathcal{F}_i$ are defined as follows: 
\begin{align*}
\mathcal{F}_1 &=\{f\} &
\mathcal{F}_2 &=\{g\} &
\end{align*}
Now consider natural concepts $A_1,A_2,A_3,A_4,\allowbreak B_1,\allowbreak B_2,\allowbreak B_3,\allowbreak B_4$ with the following feature assignments:
\begin{align*}
\varphi(A_1) &= \{f\} &
\varphi(A_2) &= \{g\} \\
\varphi(A_3) &= \{f\} &
\varphi(A_4) &= \{g\} \\
\varphi(B_1) &= \emptyset &
\varphi(B_2) &= \emptyset \\
\varphi(B_3) &= \{f\} &
\varphi(B_4) &= \{f\}
\end{align*}
Furthermore, suppose $\anaEquiv= \{1,2\} \times \{1,2\}$ and $\sigma_{12}(f)=g$. It is easy to verify that, apart from $\ana{B_1}{B_3}{B_2}{B_4}$, the conditions from Proposition \ref{propSubsumptionExtrapolation1} (i.e.\ the assertions in $K$) are all satisfied, whereas the conclusion $A_4\sqsubseteq B_4$ is not satisfied.
\end{example}

\subsection*{Counterexamples for Section \ref{secWeakSemantics}}

\noindent The next two examples show that $C$-transitivity is not satisfied under the weak semantics, for standard analogy assertions and strong analogy assertions respectively.

\begin{example}
Let $\Imf= ( \Imc,  [\mathcal{F}_1,...,\mathcal{F}_k], \mathcal{X}, \pi, \anaEquiv, \mathcal{S})$ be a weak domain constrained interpretation, where $\mathcal{X}=\{\mathcal{F}\}$, $k=4$ and the sets of features $\mathcal{F}_i$ are defined as follows: 
\begin{align*}
\mathcal{F}_1 &=\{a_1,a_1'\} &
\mathcal{F}_2 &=\{a_2,a_2'\} \\
\mathcal{F}_3 &=\{a_3,a_3'\} &
\mathcal{F}_4 &=\{a_4,a_4'\}
\end{align*}
Now consider natural concepts $A_1,A_2,A_3,B_1,B_2,B_3$ with the following feature assignments:
\begin{align*}
\varphi(A_1) &= \{a_1,a_2\} &
\varphi(A_2) &= \{a_3,a_4\} &
\varphi(A_3) &= \{a_1',a_2'\} \\
\varphi(A_4) &= \{a_3,a_4\} &
\varphi(A_5) &= \{a'_1,a'_2\} &
\varphi(A_6) &= \{a_3',a_4'\}
\end{align*}
Furthermore, suppose $\anaEquiv= \{1,2,3,4\} \times \{1,2,3,4\}$. The different bijections $\sigma_{(i,j)}$ are defined as follows:
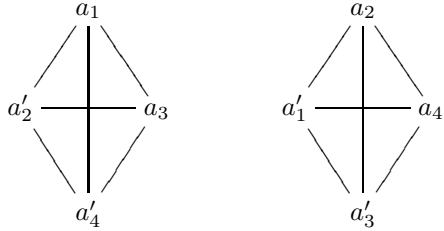
\begin{align*}
\xymatrix@C=1em{
&a_1\ar@{-}[dd]\ar@{-}[dl]\ar@{-}[dr]& \\
a_2'\ar@{-}[rr]&&a_3\\
&a_4'\ar@{-}[ul]\ar@{-}[ur]&
}
&&
\xymatrix@C=1em{
&a_2\ar@{-}[dd]\ar@{-}[dl]\ar@{-}[dr]& \\
a_1'\ar@{-}[rr]&&a_4\\
&a_3'\ar@{-}[ul]\ar@{-}[ur]&
}
\end{align*}
In these diagrams, features from different domains $\mathcal{F}_i$ and $\mathcal{F}_j$  are connected if they are mapped to each other by the  corresponding bijections $\sigma_{(i,j)}$ and $\sigma_{(j,i)}$. For instance, according to these diagrams, we have $\sigma_{(1,4)}(a_1)=a_4'$.
We have that 
\begin{align*}
\mu(A_1,A_2) &= \{\sigma_{\{(1,3),(2,4)\}}\} &
\mu(A_2,A_3) &= \{\sigma_{\{(3,2),(4,1)\}}\}\\
\mu(A_4,A_5) &= \{\sigma_{\{(3,2),(4,1)\}}\} &
\mu(A_5,A_6) &= \{\sigma_{\{(1,3),(2,4)\}}\}
\end{align*}
from which we find that $\ana{A_2}{A_3}{A_4}{A_5}$ and $\ana{A_1}{A_2}{A_5}{A_6}$ are satisfied. On the other hand, we have:
\begin{align*}
\mu(A_1,A_3)= & \{\sigma_{\{(1,2),(2,1)\}}\} &
\mu(A_4,A_6)= & \{\sigma_{\{(3,4),(4,3)\}}\}
\end{align*}
meaning that $\ana{A_1}{A_3}{A_4}{A_6}$ is not satisfied.
\end{example}

\begin{example}
Let $\Imf= ( \Imc,  [\mathcal{F}_1,...,\mathcal{F}_k], \mathcal{X}, \pi, \anaEquiv, \mathcal{S})$ be a weak domain constrained interpretation, where $\mathcal{X}=\{\mathcal{F}\}$, $k=4$ and the sets of features $\mathcal{F}_i$ are defined as follows: 
\begin{align*}
\mathcal{F}_1 &=\{a_1\} &
\mathcal{F}_2 &=\{b_1\} &
\mathcal{F}_3 &=\{a_2\} &
\mathcal{F}_4 &=\{b_2\}
\end{align*}
Now consider natural concepts $A_1,A_2,A_3,A_4,A_5,A_6$ with the following feature assignments:
\begin{align*}
\varphi(A_1) &= \{a_1,a_2\} &
\varphi(A_2) &= \{b_1,b_2\} &
\varphi(A_3) &= \{a_1,a_2\} \\
\varphi(A_4) &= \{b_1,b_2\} &
\varphi(A_5) &= \{a_1,a_2\} &
\varphi(A_6) &= \{b_1,b_2\}
\end{align*}
Furthermore, suppose $\anaEquiv= \{1,2,3,4\} \times \{1,2,3,4\}$. The different bijections $\sigma_{(i,j)}$ are uniquely defined, given that the domains are singletons, e.g.\ we have $\sigma_{(1,2)}(a_1)=b_1$.
Then clearly we have that $\sana{A_2}{A_3}{B_1}{B_2}$ and $\sana{A_1}{A_2}{A_5}{A_6}$ are satisfied, but $\sana{A_1}{A_3}{A_4}{A_6}$ is not, as 
\begin{align*}
\mu(A_1,A_3)= & \{\sigma_{\emptyset},\sigma_{\{(1,3),(3,1)\}}\} \\
\mu(B_1,B_3)= & \{\sigma_{\emptyset},\sigma_{\{(2,4),(4,2)\}}\}
\end{align*}
\end{example}

\noindent Next, we show that \eqref{eqSTransitivityA} is no longer valid for standard analogy assertions.
\begin{example}\label{counterexampleTransitivity1}
Let $\Imf= ( \Imc,  [\mathcal{F}_1,...,\mathcal{F}_k], \mathcal{X}, \pi, \anaEquiv, \mathcal{S})$ be a weak domain constrained interpretation, where $\mathcal{X}=\{\mathcal{F}\}$, $k=4$ and the sets of features $\mathcal{F}_i$ are defined as follows: 
\begin{align*}
\mathcal{F}_1 &=\{a_1,b_1,c_1,d_1,e_1\}\\
\mathcal{F}_2 &=\{a_2,b_2,c_2,d_2,e_2\}\\
\mathcal{F}_3 &=\{a_3,b_3,c_3,d_3,e_3\}\\
\mathcal{F}_4 &=\{a_4,b_4,c_4,d_4,e_4\}
\end{align*}
Now consider natural concepts $A_1,A_2,B_1,B_2,C_1,C_2$ with the following feature assignments:
\begin{align*}
\varphi(A_1) &= \{a_1,a_2\} &
\varphi(A_2) &= \{a_3,a_4\} \\
\varphi(B_1) &= \{b_1,b_2\} &
\varphi(B_2) &= \{b_3,b_4\} \\
\varphi(C_1) &= \{c_1,c_2\} &
\varphi(C_2) &= \{c_3,c_4\}
\end{align*}
Furthermore, suppose $\anaEquiv= \{1,2,3,4\} \times \{1,2,3,4\}$. The different bijections $\sigma_{(i,j)}$ are defined by the following diagrams:
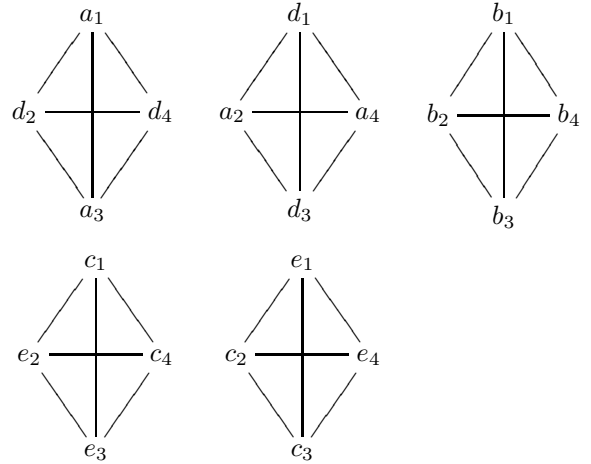
\begin{align*}
\xymatrix@C=1em{
&a_1\ar@{-}[dd]\ar@{-}[dl]\ar@{-}[dr]& \\
d_2\ar@{-}[rr]&&d_4\\
&a_3\ar@{-}[ul]\ar@{-}[ur]&
}
&&
\xymatrix@C=1em{
&d_1\ar@{-}[dd]\ar@{-}[dl]\ar@{-}[dr]& \\
a_2\ar@{-}[rr]&&a_4\\
&d_3\ar@{-}[ul]\ar@{-}[ur]&
}
&&
\xymatrix@C=1em{
&b_1\ar@{-}[dd]\ar@{-}[dl]\ar@{-}[dr]& \\
b_2\ar@{-}[rr]&&b_4\\
&b_3\ar@{-}[ul]\ar@{-}[ur]&
}
\\
\xymatrix@C=1em{
&c_1\ar@{-}[dd]\ar@{-}[dl]\ar@{-}[dr]& \\
e_2\ar@{-}[rr]&&c_4\\
&e_3\ar@{-}[ul]\ar@{-}[ur]&
}
&&
\xymatrix@C=1em{
&e_1\ar@{-}[dd]\ar@{-}[dl]\ar@{-}[dr]& \\
c_2\ar@{-}[rr]&&e_4\\
&c_3\ar@{-}[ul]\ar@{-}[ur]&
}
\end{align*}
It can now readily be verified that this interpretation satisfies $\ana{A_1}{A_2}{B_1}{B_2}$, which is witnessed by the domain translation $\sigma_{\{(1,3),(2,4)\}}$, and that this interpretation also satisfies $\ana{B_1}{B_2}{C_1}{C_2}$, which is witnessed by the domain translation $\sigma_{\{(1,4),(2,3)\}}$. But there is no domain translation that can witness $\ana{A_1}{A_2}{C_1}{C_2}$.
\end{example}
 
\noindent Next we show that Proposition \ref{propLiftingExistential} is not valid for strong analogy assertions.
\begin{example}
Let $\Imf= ( \Imc,  [\mathcal{F}_1,...,\mathcal{F}_k], \mathcal{X}, \pi, \anaEquiv, \mathcal{S})$ be a weak domain constrained interpretation, where $\mathcal{X}=\{\mathcal{F}\}$, $k=4$ and the sets of features $\mathcal{F}_i$ are defined as follows: 
\begin{align*}
\mathcal{F}_1 &=\{a_1,b_1,c_1,d_1,e_1\} &
\mathcal{F}_2 &=\{a_2,b_2,c_2,d_2,e_2\} \\
\mathcal{F}_3 &=\{a_3,b_3,c_3,d_3,e_3\} &
\mathcal{F}_4 &=\{a_4,b_4,c_4,d_4,e_4\}
\end{align*}
Now consider natural concepts $A_1,A_2,B_1,B_2$ with the following feature assignments:
\begin{align*}
\varphi(A_1) &= \{a_1,a_2\} &
\varphi(A_2) &= \{a_3,a_4\} \\
\varphi(B_1) &= \{b_1,b_2\} &
\varphi(B_2) &= \{b_3,b_4\} 
\end{align*}
Furthermore, let $\kappa_r$ be defined such that
\begin{align*}
\kappa_r(A_1)&=\{c_1,c_2\} &
\kappa_r(A_2)&=\{c_3,c_4\} \\
\kappa_r(B_1)&=\{d_1,d_2\} &
\kappa_r(B_2)&=\{d_3,d_4\}
\end{align*}
Furthermore, suppose $\anaEquiv= \{1,2,3,4\} \times \{1,2,3,4\}$. The different bijections $\sigma_{(i,j)}$ are defined as follows:
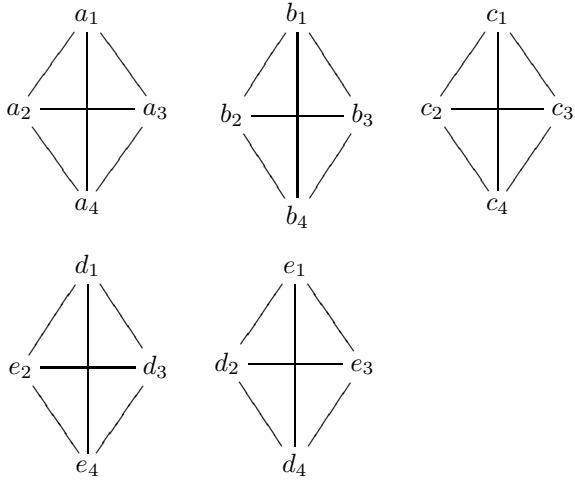
\begin{align*}
\xymatrix@C=1em{
&a_1\ar@{-}[dd]\ar@{-}[dl]\ar@{-}[dr]& \\
a_2\ar@{-}[rr]&&a_3\\
&a_4\ar@{-}[ul]\ar@{-}[ur]&
}
&&
\xymatrix@C=1em{
&b_1\ar@{-}[dd]\ar@{-}[dl]\ar@{-}[dr]& \\
b_2\ar@{-}[rr]&&b_3\\
&b_4\ar@{-}[ul]\ar@{-}[ur]&
}
&&
\xymatrix@C=1em{
&c_1\ar@{-}[dd]\ar@{-}[dl]\ar@{-}[dr]& \\
c_2\ar@{-}[rr]&&c_3\\
&c_4\ar@{-}[ul]\ar@{-}[ur]&
}
\\
\xymatrix@C=1em{
&d_1\ar@{-}[dd]\ar@{-}[dl]\ar@{-}[dr]& \\
e_2\ar@{-}[rr]&&d_3\\
&e_4\ar@{-}[ul]\ar@{-}[ur]&
}
&&
\xymatrix@C=1em{
&e_1\ar@{-}[dd]\ar@{-}[dl]\ar@{-}[dr]& \\
d_2\ar@{-}[rr]&&e_3\\
&d_4\ar@{-}[ul]\ar@{-}[ur]&
}
\end{align*}
Then we have that
\begin{align*}
\mu(A_1,A_2) &= \mu(B_1,B_2) = \{\sigma_{\{(1,3),(2,4)\}},\sigma_{\{(1,4),(2,3)\}}\}\\
\mu(\exists r.A_1,\exists r.A_2) &= \{\sigma_{\{(1,3),(2,4)\}},\sigma_{\{(1,4),(2,3)\}}\}\\
\mu(\exists r.B_1,\exists r.B_2) &= \{\sigma_{\{(1,3),(2,4)\}}\}
\end{align*}
from which we find that $\sana{A_1}{A_2}{B_1}{B_1}$ is satisfied while $\sana{(\exists 
r.A_1)}{(\exists 
r.A_2)}{(\exists 
r.B_1)}{(\exists 
r.B_2)}$ and $\sana{A_1}{A_2}{(\exists 
r.B_1)}{(\exists 
r.B_2)}$ are not.
\end{example}

\noindent The next example clarifies why Proposition \ref{propLiftingConjunction} is not satisfied under the weak semantics, neither for standard nor for strong analogy assertions.
\begin{example}
Let $\Imf= ( \Imc,  [\mathcal{F}_1,...,\mathcal{F}_k], \mathcal{X}, \pi, \anaEquiv, \mathcal{S})$ be a weak domain constrained interpretation, where $\mathcal{X}=\{\mathcal{F}\}$, $k=2$ and the sets of features $\mathcal{F}_i$ are defined as follows: 
\begin{align*}
\mathcal{F}_1 &=\{f\} &
\mathcal{F}_2 &=\{g\} 
\end{align*}
Now consider natural concepts $A_1,\allowbreak A_2,\allowbreak A_3,\allowbreak A_4,\allowbreak B_1,\allowbreak B_2,\allowbreak B_3,\allowbreak B_4$ with the following feature assignments:
\begin{align*}
\varphi(A_1) &= \{f\} &
\varphi(A_2) &= \{g\} \\
\varphi(A_3) &= \{f\} &
\varphi(A_4) &= \{g\} \\
\varphi(B_1) &= \{f\} &
\varphi(B_2) &= \{f\} \\
\varphi(B_3) &= \{f\} &
\varphi(B_4) &= \{f\} 
\end{align*}
Let $\anaEquiv = \{1,2\}\times \{1,2\}$ and $\sigma_{(1,2)}(f)=g$. Then it is clear that $\ana{A_1}{A_2}{A_3}{A_4}$, $\ana{B_1}{B_2}{B_3}{B_4}$, $\sana{A_1}{A_2}{A_3}{A_4}$ and $\sana{B_1}{B_2}{B_3}{B_4}$ are all satisfied, while $\ana{(A_1\sqcap B_1)}{(A_2\sqcap B_2)}{(A_3\sqcap B_3)}{(A_4\sqcap B_4)}$ and $\sana{(A_1\sqcap B_1)}{(A_2\sqcap B_2)}{(A_3\sqcap B_3)}{(A_4\sqcap B_4)}$ are not.
\end{example}

\noindent Finally we show that \eqref{propSubsumptionExtrapolation1} is not satisfied under the weak semantics.
\begin{example}
Let $\Imf= ( \Imc,  [\mathcal{F}_1,...,\mathcal{F}_k], \mathcal{X}, \pi, \anaEquiv, \mathcal{S})$ be a weak domain constrained interpretation, where $\mathcal{X}=\{\mathcal{F}\}$, $k=2$ and the sets of features $\mathcal{F}_i$ are defined as follows: 
\begin{align*}
\mathcal{F}_1 &=\{f\} &
\mathcal{F}_2 &=\{g\} 
\end{align*}
Now consider natural concepts $A_1,\allowbreak A_2,\allowbreak A_3,\allowbreak A_4,\allowbreak B_1,\allowbreak B_2,\allowbreak B_3,\allowbreak B_4$ with the following feature assignments:
\begin{align*}
\varphi(A_1) &= \{f,g\} &
\varphi(A_2) &= \{f,g\} \\
\varphi(A_3) &= \{f\} &
\varphi(A_4) &= \{f\} \\
\varphi(B_1) &= \{f\} &
\varphi(B_2) &= \{g\} \\
\varphi(B_3) &= \{f\} &
\varphi(B_4) &= \{g\} 
\end{align*}
Let $\anaEquiv = \{1,2\}\times \{1,2\}$ and $\sigma_{(1,2)}(f)=g$. Then it is clear that $\ana{A_1}{A_2}{A_3}{A_4}$, $\ana{B_1}{B_2}{B_3}{B_4}$, $\ana{B_1}{B_3}{B_2}{B_4}$, $\sana{A_1}{A_2}{A_3}{A_4}$,  $\sana{B_1}{B_2}{B_3}{B_4}$ and $\sana{B_1}{B_3}{B_2}{B_4}$ are all satisfied, as well as $A_1\sqsubseteq B_1$, $A_2\sqsubseteq B_2$ and $A_3\sqsubseteq B_3$. However, $A_4\sqsubseteq B_4$ is not satisfied.
\end{example}

\fi

\bibliographystyle{named}
\bibliography{refs}

\end{document}